\documentclass{article} % For LaTeX2e
\usepackage{iclr2027_conference,times}

\usepackage{amsmath,amsfonts,bm}

\def\eqref#1{equation~\ref{#1}}
\def\1{\bm{1}}

\DeclareMathAlphabet{\mathsfit}{\encodingdefault}{\sfdefault}{m}{sl}
\SetMathAlphabet{\mathsfit}{bold}{\encodingdefault}{\sfdefault}{bx}{n}

\usepackage{algorithm}
\usepackage{algpseudocode}
\usepackage{hyperref}
\hypersetup{pdftitle={},pdfauthor={},pdfsubject={},pdfkeywords={},pdfcreator={},pdfproducer={}}
\ifdefined\pdfinfoomitdate\pdfinfoomitdate=1\fi
\ifdefined\pdftrailerid\pdftrailerid{}\fi
\ifdefined\pdfsuppressptexinfo\pdfsuppressptexinfo=15\fi
\ifdefined{/Creator () /Producer ()}\fi
\usepackage{url}
\usepackage{booktabs}
\usepackage{graphicx}
\usepackage{array}
\usepackage[table]{xcolor}
\usepackage{xspace}

\usepackage{booktabs}
\usepackage{graphicx}
\usepackage{multirow}
\usepackage[table]{xcolor}
\usepackage{subcaption}
\definecolor{warmbg}{RGB}{250,247,238}

\definecolor{bestblue}{RGB}{76,145,184}
\definecolor{secondred}{RGB}{210,124,124}

\newcommand{\best}[1]{%
  \cellcolor{bestblue}\color{white}\textbf{#1}%
}

\newcommand{\second}[1]{%
  \cellcolor{secondred}\color{white}\textbf{#1}%
}
\newcommand{\name}[0]{EnvACE\xspace}
\newcommand{\envacelogotitle}{%
  \raisebox{-0.2\height}{\includegraphics[width=0.7cm]{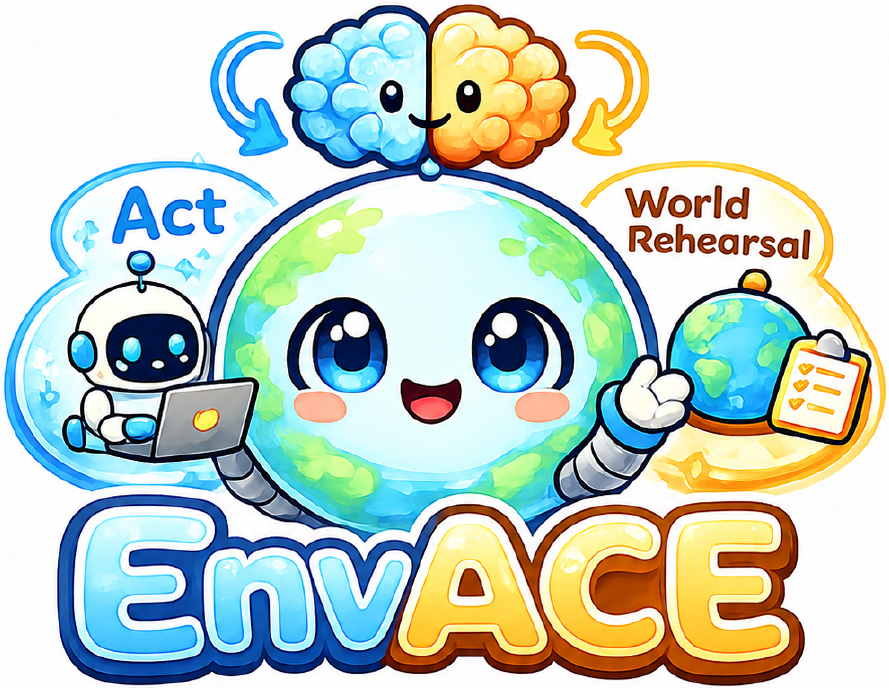}}%
}
\newcommand{\envacelogosmall}{%
  \raisebox{-0.2\height}{\includegraphics[width=0.35cm]{figure/EnvACE_logo.pdf}}%
}
\newcommand{\bestlegend}{%
  \textcolor{bestblue}{\rule{1.2ex}{1.2ex}}%
}
\newcommand{\secondlegend}{%
  \textcolor{secondred}{\rule{1.2ex}{1.2ex}}%
}
\title{
\envacelogotitle\,EnvACE: Internalizing Environment Dynamics via World Rehearsal for Agentic Reinforcement Learning
}
\author{
\\
\textbf{Zishan Xu\textsuperscript{1,7,*}},
\textbf{Zhiyuan Yao\textsuperscript{2,*}},
\textbf{Yuxin Chen\textsuperscript{3}},
\textbf{Yifu Guo\textsuperscript{4}} ,
\textbf{Zhengxi Lu\textsuperscript{2}},
\textbf{Yuquan Lu\textsuperscript{4}}, \\
\textbf{Jinyang Huang\textsuperscript{5}},
\textbf{Yan Xu\textsuperscript{7}} ,
\textbf{Yasheng Wang\textsuperscript{7,$\dagger$}},
\textbf{Weinan Zhang\textsuperscript{1}},
\textbf{Xingshan Zeng\textsuperscript{6,$\dagger$}},
\textbf{Weiwen Liu\textsuperscript{1,$\dagger$}} \\
[3mm]
$^{1}$Shanghai Jiao Tong University \quad
$^{2}$Zhejiang University \quad
$^{3}$National University of Singapore \\
$^{4}$Sun Yat-sen University \quad
$^{5}$Central South University \quad
$^{6}$The Chinese University of Hong Kong \\
$^{7}$Tencent Inc. \\
$^{*}$Equal contribution. \quad
$^{\dagger}$Corresponding authors. \\
\texttt{asheryswang@tencent.com},
\texttt{zxshamson@gmail.com},
\texttt{wwliu@sjtu.edu.cn}
}

\usepackage{xcolor}
\usepackage[most]{tcolorbox}
\usepackage{enumitem}

\definecolor{mainblue}{RGB}{65,120,210}
\definecolor{lightblue}{RGB}{242,247,255}
\iclrfinalcopy % Non-anonymous / camera-ready version.
\begin{document}

\maketitle

\begin{abstract}
Training large language model agents for long-horizon tool use typically relies on interactions with real or synthesized executable environments, whose construction and verification are costly, or on external simulators that are difficult to ground.
We introduce \envacelogosmall\,\name, an agentic reinforcement learning method that replaces external environment interaction during training with \emph{world rehearsal}.
The policy alternates between acting and rehearsal: it first generates a tool call, then plays the role of the environment to produce the response induced by that action, and conditions subsequent decisions on the rehearsed response.
Both roles are jointly optimized end-to-end using task-success rewards.
Through world rehearsal, the policy internalizes the relationship between actions and their environment responses in its parameters, yielding an agent world model that directly supports decision making.
Across BFCL-v4, $\tau^2$-Bench, VitaBench, and FinMCP-Bench, \name achieves strong and transferable performance, outperforming environment-scaling baselines in the overall evaluation.
Controlled studies further show that world rehearsal consistently improves policy learning across model scales.
At test time, the internalized world model enables private rehearsal before committed execution, yielding further gains under a moderate rehearsal budget without additional external interaction.
Our findings establish world rehearsal as a new path toward scaling LLM agent training beyond the constraints of external environments.
Our code is available at \url{https://github.com/Within-yao/EnvACE}.
% paragraph.
\end{abstract}

\section{Introduction}

\begin{figure}[htbp]
    \centering
    \includegraphics[width=1\textwidth]
    {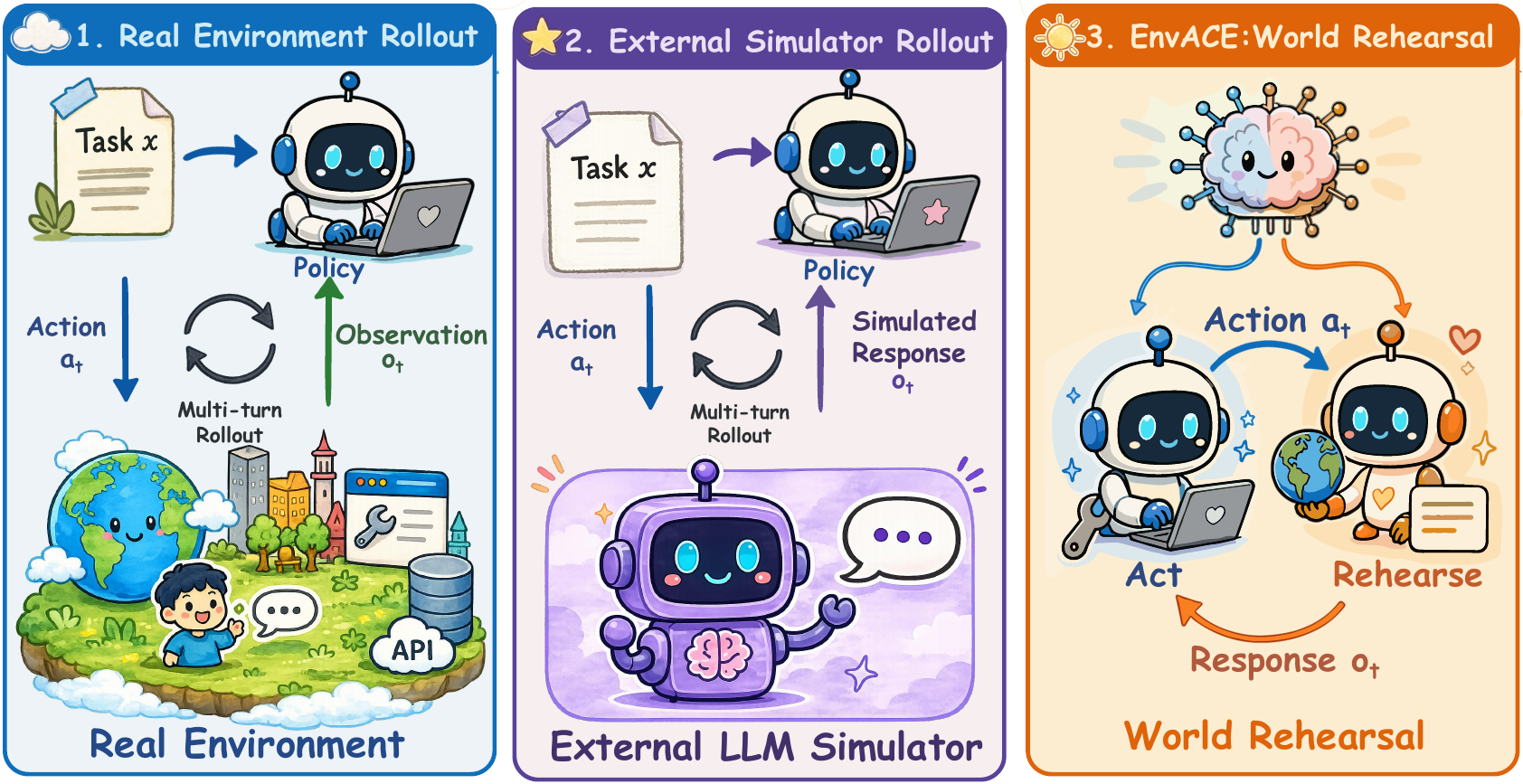}
    \caption{Comparison of three agent rollout paradigms:
real-environment rollout, external-simulator rollout, and EnvACE world rehearsal.
Existing approaches obtain environment responses from external sources,
whereas EnvACE internalizes environment responses within the policy through world rehearsal.}
    \label{fig:agent-rollout}
    \vspace{-0.5em}
\end{figure}

Large language models (LLMs) are increasingly expected to serve as agents in real-world applications, interacting with users to gather information and invoking tools to query or update the state of a specific environment~\citep{glm5.2,gpt5,kimik3}. 
Success in such settings requires effectively combining dialogue and tool use, adapting actions based on environment feedback, and operating over long interaction horizons. 
To develop such ability, we expect to expose LLMs to a sufficiently diverse range of tools and user requests while providing effective optimization signals.

To expose LLMs to rich and diverse tool-interactive environments, early work collected trajectories of tool use and user interaction for supervised fine-tuning~\citep{zeng2024agenttuning,liu2025toolace,qin2023toolllm}. 
Following the success of reinforcement learning with verifiable rewards (RLVR)~\citep{lambert2024tulu,guo2025deepseekr1}, recent work has automatically synthesized executable environments in which agents generate on-policy trajectories through interaction and learn from verifiable task rewards~\citep{dong2026agent,tu2026scaleenv},
To provide more informative training signals, some methods additionally train the policy to predict the observations resulting from its own actions as an auxiliary objective~\citep{lu2026policy,guo2025world}. 
However, scaling such environments relies on costly and complex synthesis pipelines, while their correctness becomes harder to verify as complexity grows. 
LLM-based simulators provide a less costly and more flexible alternative by directly generating environment responses, including tool outputs and user feedback, without requiring a fully executable environment~\citep{simia,ruan2024identifying}. 
However, their responses may be inaccurate or inconsistent, and grounding them still requires supervision from real environments.
Consequently, existing approaches remain fundamentally dependent on access to real environments, either for generating executable interactions or for grounding and validating simulated ones.

As illustrated in Figure~\ref{fig:agent-rollout}, existing training paradigms obtain the responses needed to continue a rollout from either an external environment or a separate simulator.
In both cases, the policy learns to act on externally supplied responses, leaving environment modeling outside the acting policy.
% Even when observation prediction is incorporated into policy training, it typically serves only as an auxiliary objective.
We argue that an effective agent policy should not only act to accomplish the task but also model how the environment responds to its actions.
Inspired by the success of learned world models~\citep{zuo2026qwen,chen2025internalizing}, we realize this idea by assigning the policy two roles: an actor that interacts with the environment and an environment that provides feedback to the actor, without external real-environment interaction. 
Rather than merely consuming or predicting externally supplied observations, the policy itself enacts the environment's responses, thereby internalizing the relationship between actions and their induced responses into its own parameters.

Toward this end, we propose EnvACE, an agentic RL method that trains a single policy to both act and rehearse its environment. 
At each turn, the policy first issues a tool call, then rehearses the environment response that the tool call would induce, and conditions its next action on this rehearsal. 
A trajectory is therefore no longer a dialogue with an external environment, but a process unfolded by the policy itself. The acting and rehearsal behaviors are jointly optimized end-to-end using a task-success reward. 
We call this process world rehearsal: the world modeling capability that prior work assigns to a separate simulator is absorbed into the policy and executed inline, enabling the policy to internalize how its actions shape environment responses while acting and world modeling unfold as a unified process.

Extensive experiments across a wide range of agentic benchmarks, including BFCL-v4, $\tau^2$-Bench, VitaBench, and FinMCP-Bench, show that EnvACE consistently outperforms strong baselines. 
Through the joint optimization of acting and rehearsal, EnvACE captures the relationship between actions and the environment responses they induce, internalizing these dynamics within the policy to support more generalizable decision making.
At test time, this internalized world model allows the policy to rehearse candidate actions before real execution, anticipate their outcomes, avoid costly mistakes, and gather useful experience without additional interaction with the external environment. 
Our experiments further examine how performance varies with the number of rehearsal attempts and show that a moderate rehearsal budget improves test-time performance.

Overall, our contributions can be concluded as follows:

\begin{itemize}
\item We introduce \textbf{world rehearsal}, in which a policy plays the role of the environment by generating the response induced by an agent action, allowing environment dynamics to be internalized without querying an external environment.
\item We propose \textbf{EnvACE}, an agentic RL method that interleaves acting with world rehearsal in self-unfolded trajectories. EnvACE uses a shared policy for both roles and jointly optimizes their behaviors end-to-end using task-success rewards.
\item We evaluate EnvACE across BFCL-v4, $\tau^2$-Bench, VitaBench, and FinMCP-Bench, where it consistently outperforms strong baselines. Controlled analyses further demonstrate the benefits of jointly learning acting and rehearsal across model scales. At test time, world rehearsal provides additional performance gains.
\end{itemize}

% Motivated by this, we introduce \textbf{EnvACE}, an agentic RL method that trains a single policy to both act and rehearse its environment. 
% Specifically, at each turn the policy first issues a tool call, then rehearses the environment's response to it, and conditions its next action on the result. 
% A trajectory is thus no longer a dialogue with an outside environment, but a process the policy unfolds on its own. 
% Learning to rehearse reshapes the policy: by modeling what its actions bring about, it internalizes the environment's dynamics and acts with a stronger sense of where those actions lead. 
% The same training therefore leaves the policy doubling as a world model of its environment, with no external simulator in the loop.
% This unlocks a form of test-time scaling unavailable to ordinary agents: before acting for real, the policy rehearses candidate actions against its own world model, anticipating their outcomes to avoid costly mistakes and gather useful experience.

% We evaluate EnvACE on a wide range of agentic benchmarks, including BFCL-v4, $\tau^2$-Bench, and VitaBench, where it delivers strong and consistent improvements. 
% We hypothesize that learning to rehearse its own interactions lets the policy better capture how its environment behaves and thereby make more generalizable decisions. 
% Under test-time scaling, world rehearsal further lets the policy play out its actions in advance to anticipate and avoid likely failures.

%contribution

\section{Related Work}

\subsection{Agentic Reinforcement Learning}

% Reinforcement learning with outcome or verifiable rewards has substantially improved language-model reasoning~\citep{guo2025deepseekr1}, with subsequent work studying adaptive rollout allocation, joint generation and verification, and tool-augmented reasoning~\citep{yao2026coba,chen2026learning,jin2025search,feng2025retool,li2025torl}. Building on interfaces that interleave reasoning, actions, and observations~\citep{yao2023react}, agentic RL extends outcome optimization across web navigation, multi-turn tool use, GUI interaction, and software engineering~\citep{qi2025webrl,singh2025agentic,lu2026ui,wei2026swe,noiserl}. The resulting long-horizon trajectories have also motivated work on training stability, credit assignment, and role-specific advantage estimation~\citep{feng2026group,feng2026dr}.

Increasingly realistic benchmarks have been developed to evaluate language agents through long-horizon interactions with real-world environments~\citep{xie2024osworld,jimenez2024swe,chen2026vitabench}. Reinforcement learning has substantially improved language-model reasoning, with recent work further advancing its efficiency and extending it to search and tool-augmented reasoning~\citep{guo2025deepseekr1,yao2026coba,jin2025search,feng2025retool,li2025torl}. Building on interfaces that interleave reasoning, actions, and observations~\citep{yao2023react}, agentic RL extends outcome optimization across web navigation, multi-turn tool use, GUI interaction, and software engineering~\citep{qi2025webrl,singh2025agentic,lu2026ui,wei2026swe,noiserl}. The resulting long-horizon trajectories have also motivated work on training stability, credit assignment, and role-specific advantage estimation~\citep{feng2026group,feng2026dr,lu2026sdar}.

\subsection{Environment Modeling for LLM Agents}

Interactive agent training requires an environment that maps actions to responses and task outcomes. Fixed executable environments provide grounded transitions but are costly to construct and scale~\citep{lu2025toolsandbox,yao2024tau}. Recent work instead synthesizes executable tools, databases, tasks, and evaluators~\citep{song2026envscaler,wang2026agent,tu2026scaleenv,dong2026agent,xu2026envfactory}, or uses language models to simulate environment feedback~\citep{simia,xiao2026webworld}. These approaches expand the available interaction experience while keeping environment dynamics external to the policy. Learned world models further support agents through lookahead and planning~\citep{chae2025web,zeng2026artis,liu2026comap}, or by incorporating environment-modeling objectives into agent training~\citep{yu2026reinforcement,lu2026policy,wang2026role,cai2026beyond}. Whereas these signals generally serve as auxiliary or planning signals, \name directly optimizes rehearsed environment responses as part of its on-policy trajectories.

\section{Preliminaries}

We formulate a tool-interactive task as a finite-horizon partially observable Markov decision process (POMDP) $\mathcal{M}=(\mathcal{S},\mathcal{A},\mathcal{O},P,\mathcal{R})$. Here, $\mathcal{S}$ is the space of environment states; $\mathcal{A}$ is the action space, including structured tool calls and final answers; $\mathcal{O}$ is the observation space, including tool outputs and user responses; $P$ specifies the environment dynamics that govern state transitions and the observation returned after an action; and $\mathcal{R}$ is the reward function that evaluates the task outcome of a completed trajectory.

Each task instance specifies an instruction $x$ and an available tool set $\mathcal{T}$. At step $t$, the policy observes the interaction history $h_t$ and generates an action $a_t\in\mathcal{A}$. For a tool call, the environment returns an observation $o_t\in\mathcal{O}$. Conditioning the response dynamics induced by $P$ on the observable interaction history gives
\begin{equation}
a_t \sim \pi_\theta(\cdot \mid h_t),\qquad
o_t \sim P(\cdot\mid h_t,a_t).
\end{equation}
The history is updated as $h_{t+1}=h_t\oplus(a_t,o_t)$, and a rollout terminates with trajectory $\tau=(x,a_1,o_1,\ldots,a_T,o_T)$.

Upon termination, the trajectory receives a scalar reward $R(\tau)$ that evaluates task performance. The policy is optimized to maximize
\begin{equation}
\max_{\theta}\ \mathcal{J}(\theta)
=
\mathbb{E}_{\tau\sim(\pi_\theta,P)}[R(\tau)].
\end{equation}
We optimize this objective with Group Relative Policy Optimization (GRPO)~\citep{guo2025deepseekr1}, which estimates advantages by normalizing each trajectory reward against other rollouts sampled for the same instruction.

This formulation exposes the conventional interaction boundary: the policy produces actions, whereas the environment supplies the observations that condition subsequent decisions. EnvACE revises this boundary by assigning observation generation itself to a rehearsal role of the policy.

\section{Method}
\label{sec:method}

\begin{figure*}[t]
\centering
\includegraphics[width=\textwidth]{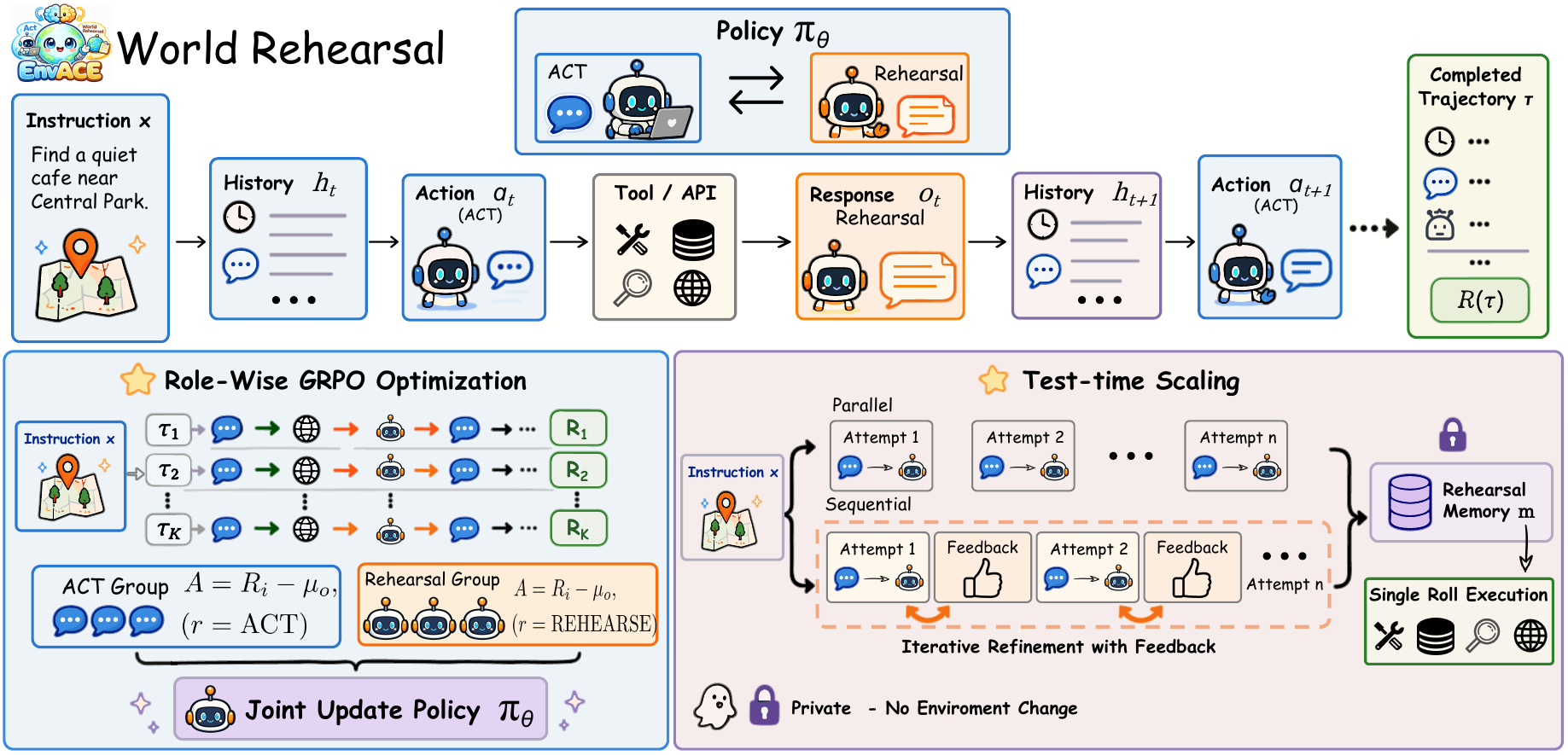}
\caption{
\textbf{Overview of EnvACE and world rehearsal.}
Unlike conventional agentic RL, where an external environment provides observations after each action, EnvACE internalizes the interaction loop into a single policy.
At each turn, the policy first produces an action, then rehearses the corresponding environment response, and conditions its next decision on the self-generated observation.
This unified act--rehearse process enables the policy to learn environment dynamics during training and to simulate multiple candidate actions at test time before executing the most promising one in the real environment.
}
\label{fig:envace_method}
\end{figure*}

Figure~\ref{fig:envace_method} presents an overview of EnvACE, which comprises world rehearsal, role-wise GRPO optimization, and test-time scaling. During training, the policy alternates between acting and rehearsal, appending each generated environment response to the interaction history so that the trajectory unfolds without an external environment. Role-wise GRPO uses separate baselines for the two roles while jointly updating the policy, enabling it to internalize how actions shape environment responses. At test time, the policy performs private rehearsals in parallel or sequentially, summarizes them into a rehearsal memory, and uses this memory to guide a committed execution in the external environment.

\subsection{World Rehearsal}

% EnvACE turns environment response generation into an explicit role of the policy. We instantiate a role-conditioned policy with an acting role and a rehearsal role. The acting role first produces an environment-facing action from the current history:
% \begin{equation}
% a_t \sim \pi_\theta(\cdot \mid h_t, \textsc{Act}).
% \end{equation}
% EnvACE then invokes the same policy under the rehearsal role, conditioning it on the current history and the proposed action to generate the environment response that this action would induce:
% \begin{equation}
% \hat{o}_t \sim \pi_\theta(\cdot \mid h_t, a_t, \textsc{Rehearse}).
% \end{equation}
% The generated response is committed into the trajectory as the next observation, and the acting role continues from the resulting history.

% The key distinction is that $\hat{o}_t$ is not merely a prediction target; it becomes the state that conditions later decisions in the same rollout. Thus acting and rehearsal are optimized together through downstream trajectory outcomes, pressuring the same parameters to learn both which actions to take and what those actions make happen. Over repeated rollouts, these action--consequence regularities are absorbed into $\theta$, allowing the policy to internalize environment dynamics as an agent world model.

A conventional agent rollout can proceed only when an external environment or simulator provides a response after each action. Consequently, the scalability and quality of policy learning remain tied to an external response provider, while the ability to model environment responses remains outside the acting policy. World rehearsal changes this division of labor by making environment response generation an explicit role of the policy itself. This allows the policy to unfold its own training trajectories without querying an external environment.

Concretely, EnvACE assigns two roles to a shared policy $\pi_\theta$: an acting role and a rehearsal role. Given the interaction history $h_t$, the acting role first generates an environment-facing action:
\begin{equation}
a_t \sim \pi_\theta(\cdot \mid h_t, \textsc{Act}).
\end{equation}
Conditioned on the history and the generated action, the rehearsal role then generates the corresponding environment response:
\begin{equation}
\hat{o}_t \sim \pi_\theta(\cdot \mid h_t, a_t, \textsc{Rehearse}).
\end{equation}
The generated response is appended to the interaction history,
\begin{equation}
h_{t+1}=h_t\oplus(a_t,\hat{o}_t),
\end{equation}
from which the acting role makes its next decision. Acting and rehearsal alternate in this manner until termination, so subsequent actions are conditioned on the environment responses generated by the policy itself. Through repeated rehearsal, the relationship between actions and their environment responses is absorbed into $\theta$, enabling the policy to internalize environment dynamics as an agent world model.

\subsection{Role-Wise GRPO Optimization}

For each instruction $x$, EnvACE samples a group of $K$ rollouts $\{\tau_i\}_{i=1}^{K}$, each receiving a trajectory-level reward $R_i=R(\tau_i)$. Depending on the task, this reward can be provided by either a verifiable outcome evaluator or a checklist-based LLM judge.
Let $\mathcal{Y}_i=\{y_{i,m}\}_{m=1}^{M_i}$ denote all policy outputs generated in rollout $\tau_i$, where each output is associated with a role $r_{i,m}\in\{\textsc{Act},\textsc{Rehearse}\}$. Every policy output in the same rollout inherits its trajectory reward $R_i$.

For each role $r$, we collect all policy outputs generated under that role across the $K$ rollouts:
\begin{equation}
\mathcal{G}_{x,r}
=
\left\{
y_{i,m}
\mid
i=1,\ldots,K,\;
r_{i,m}=r
\right\}.
\end{equation}
EnvACE computes a separate reward baseline for each role and defines the role-wise advantage $A_{i,m}$ of output $y_{i,m}$ relative to this baseline:
\begin{equation}
\mu_{x,r}
=
\frac{1}{|\mathcal{G}_{x,r}|}
\sum_{y_{j,n}\in\mathcal{G}_{x,r}}R_j,
\qquad
A_{i,m}
=
R_i-\mu_{x,r_{i,m}}.
\end{equation}
Thus, outputs from the same trajectory receive the same reward, while their advantages are computed relative to other outputs generated under the same role.

The shared policy is optimized using the clipped GRPO objective:
\begin{equation}
\max_{\theta}\ \mathcal{J}(\theta)
=
\mathbb{E}_{x,i,m,\ell}
\left[
\min\left(
\rho_{i,m,\ell}(\theta)A_{i,m},
\operatorname{clip}\left(
\rho_{i,m,\ell}(\theta),
1-\epsilon,1+\epsilon
\right)A_{i,m}
\right)
\right],
\end{equation}
where $\rho_{i,m,\ell}(\theta)$ is the standard GRPO likelihood ratio for the $\ell$-th token of $y_{i,m}$. Although the reward baselines are computed separately for the two roles, outputs from both roles jointly update the shared policy parameters $\theta$.

\subsection{Test-Time Scaling}
\label{sec:test_time_scaling}
After training, EnvACE uses world rehearsal to scale inference-time computation before interacting with the external environment. Let $\Pi_\theta$ denote the rollout distribution induced by the act--rehearse loop. Given a new instruction $x$, EnvACE performs $N$ private rehearsal attempts, each starting from the same initial task context. An attempt produces an imagined trajectory $\tilde{\tau}^{(n)}$, which the policy then evaluates to obtain feedback $f^{(n)}$ comprising an assessment and a suggestion for revision.

We consider two modes that differ in how information is shared across rehearsal attempts:
\begin{equation}
\begin{aligned}
\textsc{Parallel:}\quad
&\tilde{\tau}^{(n)}
\sim
\Pi_\theta(\cdot\mid x),
\\
\textsc{Sequential:}\quad
&\tilde{\tau}^{(n)}
\sim
\Pi_\theta\left(
\cdot\mid
x,\{(\tilde{\tau}^{(j)},f^{(j)})\}_{j<n}
\right).
\end{aligned}
\end{equation}
In the parallel mode, all attempts are generated independently from the same context; their feedback is retained for final aggregation but is not exposed to other attempts. In the sequential mode, each new attempt observes the previous rehearsal trajectories together with their assessments and revision suggestions, allowing it to refine earlier decisions and avoid repeated failures.

After completing the rehearsal attempts, EnvACE summarizes all trajectories and self-evaluations into a compact rehearsal memory $m_x$. The acting role conditions on $m_x$ during a single committed execution in the external environment. The rehearsals themselves remain private and do not alter the external environment.
\section{Experiments}

\subsection{Experimental Setup}
\label{sec:setup}

\paragraph{Benchmarks}

We evaluate \name on four complementary agentic benchmarks: BFCL-v4~\citep{patil2025bfcl}, $\tau^2$-Bench~\citep{barres2025tau2}, VitaBench~\citep{he2025vitabench}, and FinMCP-Bench~\citep{zhu2026finmcp}. BFCL-v4 evaluates function calling and tool use across single-turn, multi-turn, and agentic tasks. $\tau^2$-Bench and VitaBench evaluate agents in stateful, realistic service environments: the former covers Retail, Telecom, and Airline, while the latter includes food delivery, in-store consumption, online travel, and cross-domain scenarios. FinMCP-Bench evaluates financial agents on real-world tool-use tasks through the Model Context Protocol (MCP). We follow the official evaluation protocol of each benchmark.

\paragraph{Baselines}
We compare \name against Qwen3 models at different scales (1.7B, 4B, and 8B), as well as Qwen3 models trained with standard GRPO. We also include representative tool-use training methods based on environment simulation or synthesis. Simulator-8B~\citep{simia} uses reasoning models to simulate environment feedback, while TOUCAN-7B~\citep{xu2025toucan} synthesizes large-scale trajectories from real-world MCP environments. EnvScaler-8B~\citep{song2026envscaler} programmatically synthesizes tool-interactive environments, AWM-8B/14B~\citep{wang2026agent} builds code-driven environments with database-backed states, and ScaleEnv-8B~\citep{tu2026scaleenv} constructs interactive environments and verifiable tasks from scratch.

\paragraph{Implementation Details}
% We conduct our main experiments with Qwen3-8B~\citep{yang2025qwen3} and train on the data introduced by CM2~\citep{cm2}. We optimize the model for one epoch with a learning rate of $1\times10^{-6}$, a batch size of 16, and four rollouts per prompt. The KL coefficient is set to $1\times10^{-4}$, and the maximum rollout prompt and response lengths are 12,000 and 8,000 tokens, respectively. We use Qwen3-30B-A3B as the LLM judge during training and report Avg@4 results over four runs. Training is implemented with the verl framework and conducted on 16 NVIDIA H20 GPUs.
We conduct experiments with Qwen3 backbones at three model scales: 1.7B, 4B, and 8B~\citep{yang2025qwen3}. All models are trained on the dataset introduced by CM2~\citep{cm2}. We optimize the model for 470 training steps with a learning rate of $1\times10^{-6}$, a batch size of 16, and four rollouts per prompt. The KL coefficient is set to $1\times10^{-4}$, and the entropy coefficient is set to 0.0. At each training step, we sample 64 instances. The maximum input and response lengths are set to 12,000 and 8,000 tokens, respectively, and each agent trajectory is allowed up to 30 interaction turns. During training, we use Qwen3-30B-A3B as the LLM judge. For non-TTS experiments, we report Avg@4 results averaged over four independent runs. Due to the high computational cost of TTS experiments, we report results from a single run only. For the experiments in Table~\ref{tab:tts_scaling_results}, both the acting role and the rehearsal role use a sampling temperature of 1.0 and a top-$p$ of 1.0 to encourage reasoning diversity and broader exploration of the action space. For the remaining roles, we use a lower temperature of 0.01 to improve robustness and reproducibility. For all other experiments outside Table~\ref{tab:tts_scaling_results}, the acting role uses a sampling temperature of 0.6 and a top-$p$ of 0.95. Training is implemented with the verl framework and conducted on 16 NVIDIA H20 GPUs.

% Detailed training configurations are provided in the Appendix~\ref{sec:Training and Evaluation Details}.

\subsection{Main Results}

\begin{table*}[t]
\centering
\tiny
\setlength{\tabcolsep}{2.2pt}
\renewcommand{\arraystretch}{1.08}

\resizebox{\textwidth}{!}{%
\begin{tabular}{lccccccc|cccc|ccccc|c}
\toprule

\multirow{2}{*}{\textbf{Method}}
& \multicolumn{7}{c|}{\textbf{BFCL V4}}
& \multicolumn{4}{c|}{\textbf{$\tau^2$-Bench}}
& \multicolumn{5}{c|}{\textbf{VitaBench}}
& \multirow{2}{*}{\textbf{Overall}} \\

\cmidrule(lr){2-8}
\cmidrule(lr){9-12}
\cmidrule(lr){13-17}

& Web & Mem. & Multi & NoLive & Live & Irrel. & Avg.
& Retail & Telecom & Airline & Avg.
& Cross & Deliv. & Inst. & OTA & Avg.
& \\

\midrule

\rowcolor{warmbg}
\multicolumn{18}{c}{%
\textbf{Open-Source Foundation Models (1.7B--8B)}} \\
\midrule

Qwen3-1.7B
& 1.88
& 12.58
& 15.75
& 82.94
& 74.06
& 75.71
& 30.89
& 4.6
& 11.6
& 15.0
& 10.4
& 0.0
& 3.7
& 3.3
& 0.7
& 1.9
& 14.41 \\

Qwen3-4B
& 8.25
& 20.86
& 37.25
& 88.16
& 81.55
& 78.33
& 41.80
& 32.7
& 13.2
& 38.0
& 27.9
& 1.8
& 13.7
& 18.2
& 4.8
& 9.6
& 26.43 \\

Qwen3-8B
& 10.63
& 22.15
& 42.09
& 88.09
& 80.90
& 79.58
& 44.04
& 41.2
& \second{20.8}
& 28.0
& 30.0
& 1.2
& 15.5
& 24.8
& 4.0
& 11.4
& 28.48 \\

\midrule

\rowcolor{warmbg}
\multicolumn{18}{c}{%
\textbf{Open-Source Environment Scaling Methods (7B--14B)}} \\
\midrule

Simulator-8B
& 9.38
& 5.65
& 1.47
& 32.46
& 44.30
& \best{86.54}
& 19.78
& \second{49.3}
& 15.1
& \best{51.0}
& \best{38.5}
& 0.0
& 1.2
& 4.8
& 1.0
& 1.8
& 20.03 \\

TOUCAN-7B
& 14.00
& 15.97
& 22.59
& 76.11
& 72.61
& 76.89
& 35.33
& 30.0
& 10.1
& 27.0
& 22.4
& 0.5
& 6.9
& 3.0
& 0.8
& 2.8
& 20.18 \\

EnvScaler-8B
& \best{15.38}
& 19.52
& \best{52.25}
& 86.40
& \second{81.70}
& 76.03
& \second{47.07}
& 48.0
& 16.2
& 34.5
& 32.9
& \second{6.0}
& 20.8
& \second{29.5}
& \second{7.0}
& 15.8
& 31.92 \\

AWM-8B
& \second{15.00}
& 19.46
& 42.81
& \second{88.47}
& 80.64
& 76.41
& 44.29
& 41.3
& 18.2
& 34.0
& 31.2
& 1.5
& 16.0
& 19.2
& 4.2
& 10.2
& 28.56 \\

AWM-14B
& 14.88
& 21.88
& \second{49.06}
& \best{89.46}
& 81.41
& 81.59
& \best{47.32}
& 41.9
& 19.1
& 31.0
& 30.7
& \best{8.0}
& \second{24.8}
& \best{33.8}
& \best{11.8}
& \best{19.6}
& \second{32.54} \\

ScaleEnv-8B
& -- & -- & -- & -- & -- & -- & --
& \best{50.9}
& \best{27.2}
& 37.5
& \best{38.5}
& 3.0
& \best{26.3}
& 23.8
& \second{7.0}
& 15.0
& -- \\

\midrule

\rowcolor{warmbg}
\name-1.7B
& 3.75
& 15.70
& 14.38
& 83.13
& 74.76
& 78.14
& 31.81
& 7.9
& 14.5
& 23.5
& 15.3
& 0.0
& 7.5
& 5.0
& 0.3
& 3.2
& 16.77 \\

\rowcolor{warmbg}
\name-4B
& 12.75
& \best{24.74}
& 29.63
& 87.86
& \best{82.27}
& \second{84.79}
& 41.88
& 36.8
& 12.7
& 36.0
& 28.5
& 1.5
& 18.0
& 19.0
& 4.5
& 10.8
& 27.06 \\

\rowcolor{warmbg}
\name-8B
& 12.25
& \second{24.03}
& 45.29
& 87.59
& 81.20
& 83.19
& 46.04
& 48.9
& 17.3
& \second{44.0}
& \second{36.7}
& \second{6.0}
& 24.0
& 27.0
& \second{7.0}
& \second{16.0}
& \best{32.91} \\

\bottomrule
\end{tabular}%
}

\caption{
Benchmark results across BFCL V4, $\tau^2$-Bench, and VitaBench.
Overall is the arithmetic mean of the BFCL V4 Avg.,
$\tau^2$-Bench Avg., and VitaBench Avg.
\bestlegend{} Blue cells indicate the highest result in each column,
while \secondlegend{} red cells indicate the second-highest distinct
result. 
}
\label{tab:bfcl_tau2_vitabench_results}
\end{table*}

% \begin{table}[t]
% \centering
% \small
% \setlength{\tabcolsep}{6pt}
% \renewcommand{\arraystretch}{1.05}

% \begin{tabular}{lccc}
% \toprule
% Method & TR & TP & TF1 \\
% \midrule

% Qwen3-8B
% & 0.4318 & 0.3947 & 0.4124 \\

% AWM-8B
% & 0.4643 & 0.3918 & 0.4250 \\

% Simulator-8B
% & 0.1136 & 0.2672 & 0.1595 \\

% EnvScaler-8B
% & 0.4935 & 0.3918 & 0.4368 \\

% EnvACE
% & 0.4123 & 0.5404 & 0.4678 \\

% \bottomrule
% \end{tabular}

% \caption{Performance comparison on FinMCP-Bench using TR, TP, and TF1 metrics.}
% \label{tab:tr_tp_tf1_results}
% \end{table}

% \begin{table}[t]
% \centering
% \small
% \setlength{\tabcolsep}{6pt}
% \renewcommand{\arraystretch}{1.05}

% \begin{tabular}{lccc}
% \toprule
% \textbf{Method} & \textbf{TR} & \textbf{TP} & \textbf{TF1} \\
% \midrule

% Qwen3-8B
% & 0.4318
% & \second{0.3947}
% & 0.4124 \\

% AWM-8B
% & \second{0.4643}
% & 0.3918
% & 0.4250 \\

% Simulator-8B
% & 0.1136
% & 0.2672
% & 0.1595 \\

% EnvScaler-8B
% & \best{0.4935}
% & 0.3918
% & \second{0.4368} \\

% \rowcolor{warmbg}
% \textbf{EnvACE-8B}
% & \textbf{0.4123}
% & \best{0.5404}
% & \best{0.4678} \\

% \bottomrule
% \end{tabular}

% \caption{
% Performance comparison on FinMCP-Bench using TR, TP, and TF1 metrics.
% \bestlegend{} Blue cells indicate the highest result in each column,
% while \secondlegend{} red cells indicate the second-highest result.
% }
% \label{tab:tr_tp_tf1_results}
% \end{table}

\begin{table}[t]
\centering
\small
\setlength{\tabcolsep}{6pt}
\renewcommand{\arraystretch}{1.05}

\begin{tabular}{lccc}
\toprule
\textbf{Method} & \textbf{TR (\%)} & \textbf{TP (\%)} & \textbf{TF1 (\%)} \\
\midrule

Qwen3-8B
& 43.18
& \second{39.47}
& 41.24 \\

AWM-8B
& \second{46.43}
& 39.18
& 42.50 \\

Simulator-8B
& 11.36
& 26.72
& 15.95 \\

EnvScaler-8B
& \best{49.35}
& 39.18
& \second{43.68} \\

\rowcolor{warmbg}
EnvACE-8B
& 41.23
& \best{54.04}
& \best{46.78} \\

\bottomrule
\end{tabular}

\caption{
Performance comparison on FinMCP-Bench using TR, TP, and TF1 metrics.
Values are reported as percentages (\%).
\bestlegend{} Blue cells indicate the highest result in each column,
while \secondlegend{} red cells indicate the second-highest result.
}
\label{tab:tr_tp_tf1_results}
\end{table}

% Table~\ref{tab:bfcl_tau2_vitabench_results} presents the main results across BFCL V4, $\tau^2$-Bench, and VitaBench. \name achieves an Overall score of 32.91\%, outperforming all environment-scaling baselines with complete results on the three benchmarks. It surpasses EnvScaler-8B and AWM-14B by 0.99\% and 0.37\%, respectively, and is only 0.17\% behind Qwen3-14B despite using an 8B backbone. This strong overall result reflects consistent performance across heterogeneous environments. On BFCL V4, \name reaches 46.04\%, outperforming Qwen3-8B and AWM-8B by 2.00\% and 1.75\%, respectively, while remaining within 1.03\% of EnvScaler-8B. On $\tau^2$-Bench, it obtains the second-highest average of 36.7\%, exceeding EnvScaler-8B, AWM-8B, and AWM-14B by 3.8\%, 5.5\%, and 6.0\%, respectively. On VitaBench, \name achieves 16.0\%, the best average among all 7B--8B methods, outperforming EnvScaler-8B and ScaleEnv-8B by 0.2\% and 1.0\%.

Table~\ref{tab:bfcl_tau2_vitabench_results} presents the main results across BFCL V4, $\tau^2$-Bench, and VitaBench. \name achieves an Overall score of 32.91\%, outperforming all environment-scaling baselines with complete results on the three benchmarks. It surpasses EnvScaler-8B and AWM-14B by 0.99\% and 0.37\%, respectively. This strong overall result reflects consistent performance across heterogeneous environments. On BFCL V4, \name reaches 46.04\%, outperforming Qwen3-8B and AWM-8B by 2.00\% and 1.75\%, respectively, while remaining within 1.03\% of EnvScaler-8B. On $\tau^2$-Bench, it obtains the second-highest average of 36.7\%, exceeding EnvScaler-8B, AWM-8B, and AWM-14B by 3.8\%, 5.5\%, and 6.0\%, respectively. On VitaBench, \name achieves 16.0\%, the best average among all 7B--8B methods, outperforming EnvScaler-8B and ScaleEnv-8B by 0.2\% and 1.0\%.

Table~\ref{tab:tr_tp_tf1_results} reports the results on FinMCP-Bench. \name achieves the best TF1 score of 46.78\%, outperforming EnvScaler-8B and AWM-8B by 3.10\% and 4.28\%, respectively. It also obtains the highest tool precision of 54.04\%. Although its tool recall is not the highest, the resulting TF1 demonstrates the strongest balance between precision and recall. Taken together, these results demonstrate that world rehearsal provides a more effective and scalable training paradigm than representative methods relying on real-environment interaction or external simulators. By jointly learning to act and generate environment responses within a single policy, \name acquires transferable tool-use capabilities across heterogeneous and specialized scenarios.

\subsection{Effectiveness of World Rehearsal}

\paragraph{World Rehearsal Improves Policy Learning}
% Figure~\ref{fig:tau2_ablation} provides a controlled comparison with standard GRPO using Qwen3-8B. \name improves the BFCL V4 average from 44.68\% to 46.04\% and the $\tau^2$-Bench average from 31.2\% to 36.7\%, corresponding to gains of 1.36\% and 5.5\%, respectively. The consistent improvements across both benchmarks show that jointly learning to act and rehearse environment responses strengthens policy learning, with a particularly pronounced benefit on the stateful interactions in $\tau^2$-Bench.

\begin{figure*}[t]
    \centering

    \begin{minipage}[t]{0.48\textwidth}
        \centering
        \includegraphics[width=\linewidth]{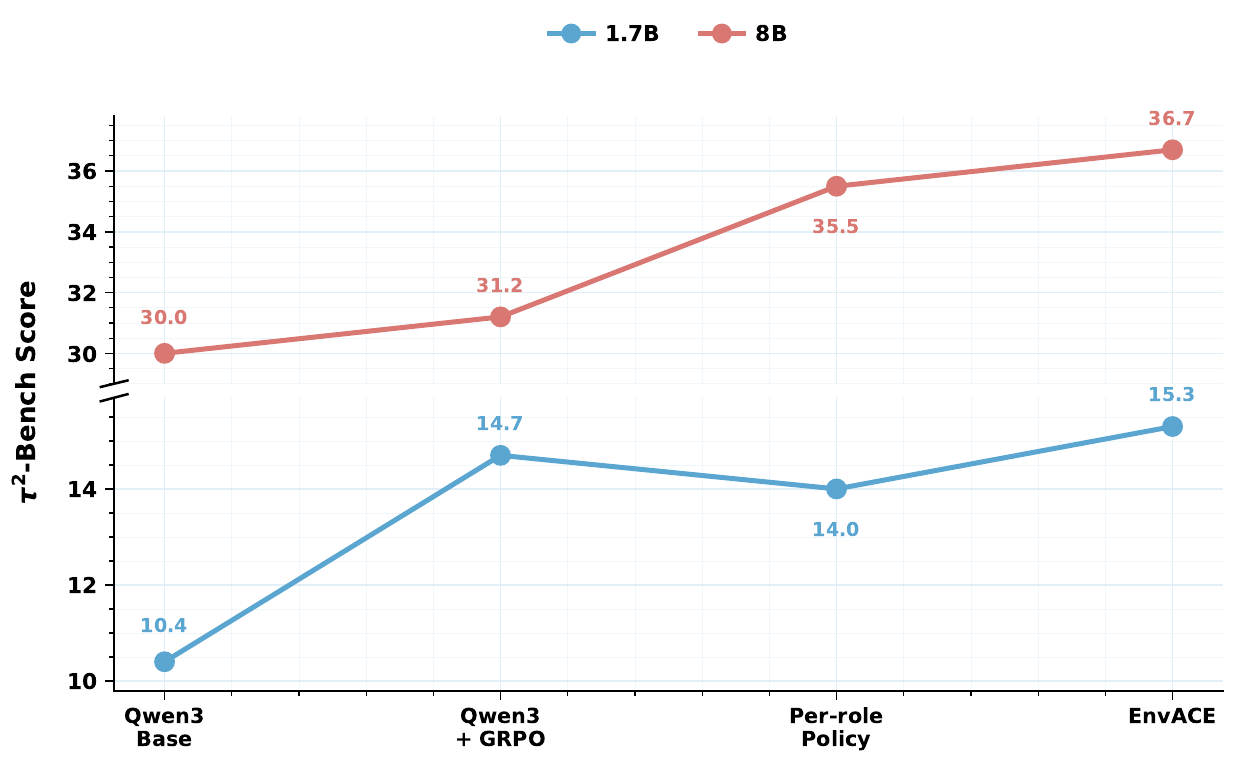}
        \caption{
        \textbf{Ablation results on $\tau^2$-Bench.}
        EnvACE achieves the best performance at both the 1.7B and 8B scales.
        }
        \label{fig:tau2_ablation}
    \end{minipage}
    \hfill
    \begin{minipage}[t]{0.48\textwidth}
        \centering
        \includegraphics[width=\linewidth]{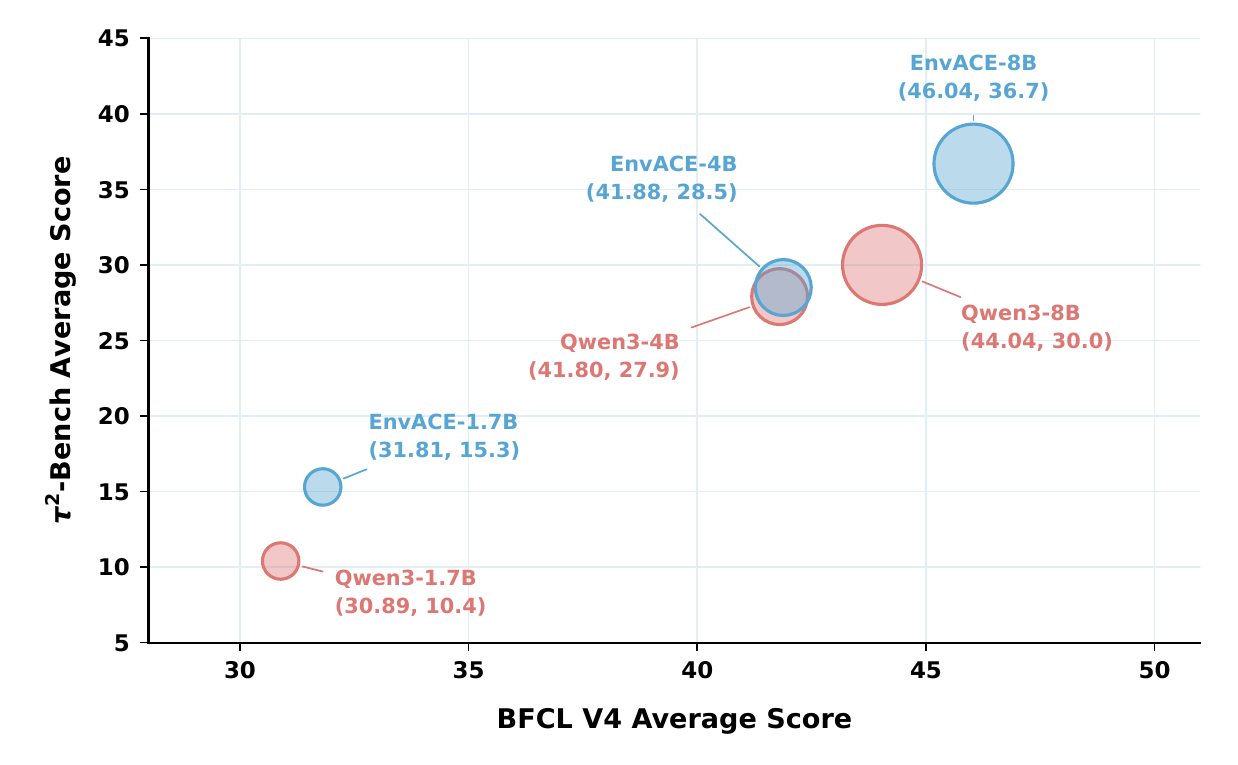}
        \caption{
        \textbf{EnvACE and base-model performance across model scales.}
        EnvACE (blue) and Qwen3 (red) at 1.7B, 4B, and 8B on both benchmarks.
        }
        \label{fig:envace_model_size}
    \end{minipage}

\end{figure*}

Figure~\ref{fig:tau2_ablation} provides a controlled comparison with standard GRPO at the 8B scale. On $\tau^2$-Bench, \name improves the average score from 31.2\% to 36.7\%, a gain of 5.5\%. This improvement is particularly relevant for stateful, multi-turn interactions, where each action depends on the environment responses observed in previous turns. By jointly learning to act and rehearse these responses, the policy can better anticipate interaction dynamics and make more effective subsequent decisions.

\paragraph{Internalizing Environment Dynamics}
% Figure~\ref{fig:tau2_ablation} also compares \name with Per-role Policy, a variant that assigns separate policies to acting and rehearsal. Both variants generate environment responses, but only \name shares parameters between the two roles. Parameter sharing improves the BFCL V4 average from 45.01\% to 46.04\% and the $\tau^2$-Bench average from 35.5\% to 36.7\%. Because acting and rehearsal update the same parameters, knowledge acquired during rehearsal about how the environment responds to actions is directly incorporated into the acting policy. These consistent gains support our hypothesis that parameter sharing enables the policy to internalize environment dynamics as an agent world model and use this knowledge for better decision making.

Figure~\ref{fig:tau2_ablation} also compares \name with Per-role Policy, a variant that assigns separate policies to acting and rehearsal. Both variants generate environment responses, but only \name shares parameters between the two roles. On $\tau^2$-Bench, parameter sharing improves the average score from 35.5\% to 36.7\%, a gain of 1.2\%. Because acting and rehearsal update the same parameters, knowledge acquired during rehearsal about how the environment responds to actions is directly incorporated into the acting policy. This gain supports our hypothesis that parameter sharing enables the policy to internalize environment dynamics as an agent world model and use this knowledge for better decision making.

\paragraph{Training Dynamics}
% Figure~\ref{fig:share8b_tau2_training_curve} further illustrates the optimization behavior of \name. As training proceeds, the policy progressively improves its task performance through world rehearsal. Performance continues to increase during the final stage and reaches its highest score of 36.7\% at step 470. This trajectory indicates that world rehearsal provides an effective learning signal throughout RL training, progressively strengthening the policy's task-solving capability.

\begin{figure}[t]
    \centering

    \begin{minipage}[t]{0.49\columnwidth}
        \centering
        \includegraphics[
            width=\linewidth
        ]{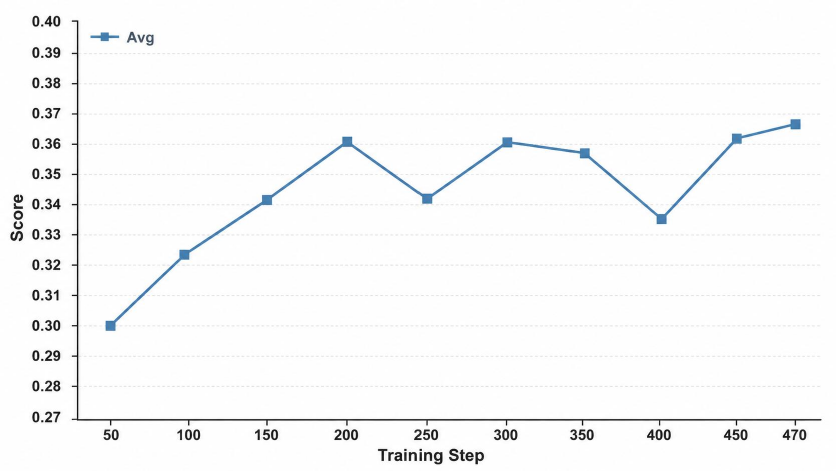}
        \captionof{figure}{
            Evaluation performance of EnvACE-8B throughout RL training on $\tau^2$-bench.
        }
        \label{fig:share8b_tau2_training_curve}
    \end{minipage}
    \hfill
    \begin{minipage}[t]{0.49\columnwidth}
        \centering
        \includegraphics[
            width=\linewidth
        ]{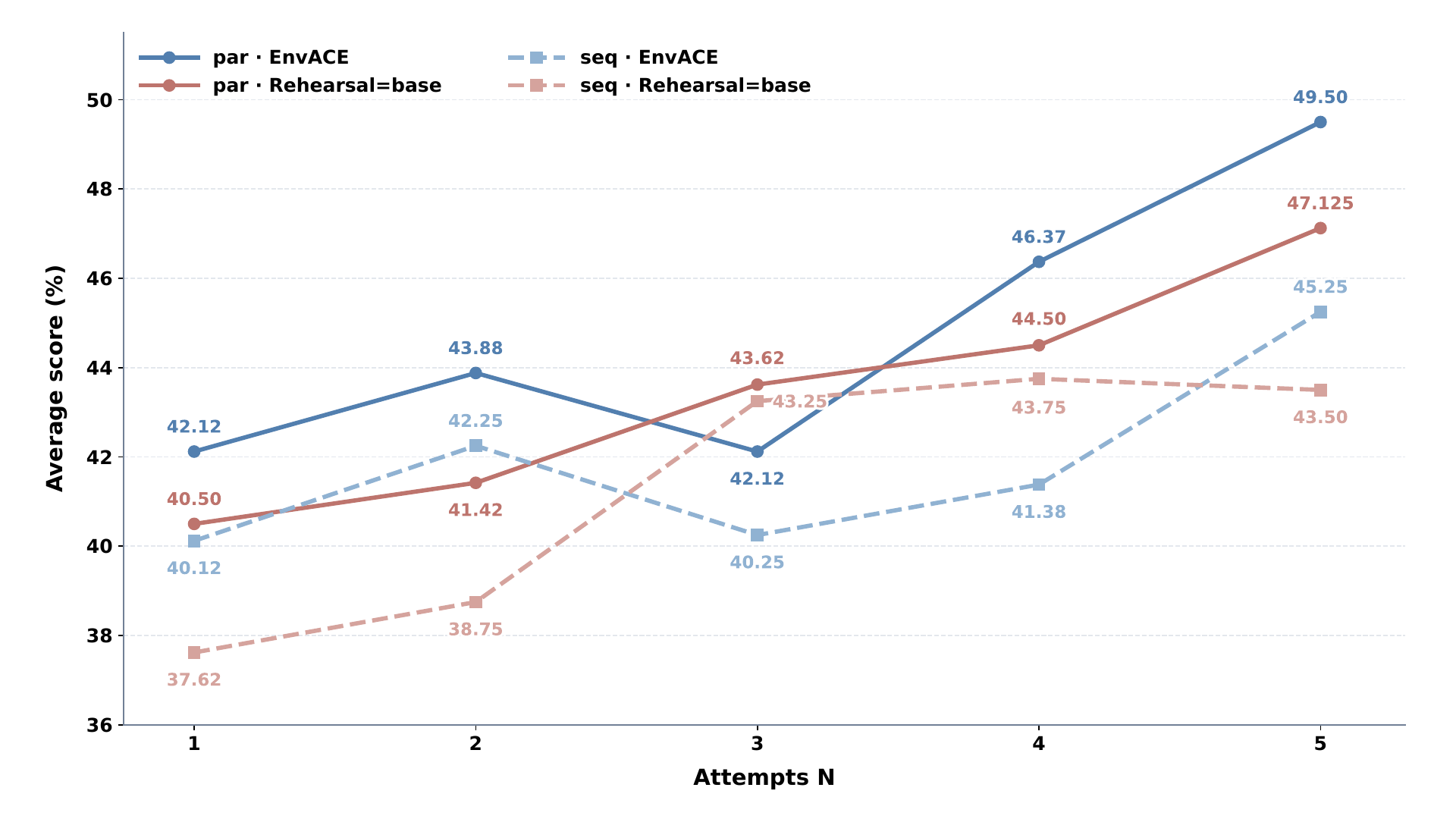}
        \captionof{figure}{
            % TTS scaling performance. $N=2$ achieves the best results, while larger $N$ introduces context-length issues.
TTS scaling performance on BFCL Multi-Turn.
        }
        \label{fig:bfcl-tts-scaling}
    \end{minipage}

\end{figure}

Figure~\ref{fig:share8b_tau2_training_curve} further illustrates the optimization behavior of \name. The offline evaluation score exhibits an overall upward trend during training, increasing from 30.0\% at step 50 to 36.7\% at step 470. Although performance fluctuates at intermediate checkpoints, it recovers in the later stage and reaches its highest observed score at the final checkpoint. This overall trend indicates that world rehearsal provides a sustained learning signal during RL training and progressively strengthens the policy's task-solving capability.

% \subsection{Effectiveness of World Rehearsal}

% \paragraph{World Rehearsal Improves Policy Learning}
% Table~\ref{tab:bfcl_tau2_results} presents a controlled ablation on Qwen3-8B. Compared with standard GRPO, \name achieves consistent improvements across BFCL V4 and $\tau^2$-Bench. The benefit is particularly pronounced on $\tau^2$-Bench, where \name reaches 36.7\%, outperforming standard GRPO by 5.5\%. This result shows that policy learning benefits from modeling the environment responses induced by its actions and conditioning subsequent decisions on those responses.

% \paragraph{Internalizing Environment Dynamics}
% We further compare \name with the variant that assigns a separate policy to each role. Although both variants explicitly model environment responses, sharing the policy improves the average score from 45.01\% to 46.04\% on BFCL V4 and from 35.5\% to 36.7\% on $\tau^2$-Bench. We attribute these gains to the internalization enabled by parameter sharing. By performing both acting and rehearsal, the shared policy learns the regularities between agent actions and the outcomes they produce, and uses this knowledge to guide subsequent decisions. Through this joint learning process, the policy develops both decision making and world modeling capabilities, enabling it to better anticipate environment transitions and solve complex tool use tasks.

\subsection{Effects across Model Scales}
We apply the same training recipe to the 1.7B, 4B, and 8B backbones to examine whether \name scales with model capacity. As shown in Figure~\ref{fig:envace_model_size}, scaling from 1.7B to 8B improves the BFCL V4 average from 31.81\% to 46.04\%, a gain of 14.23\%, and the $\tau^2$-Bench average from 15.3\% to 36.7\%, a gain of 21.4\%. Figure~\ref{fig:tau2_ablation} further shows that \name outperforms standard GRPO at both scales, with the improvement becoming more pronounced at 8B. These results demonstrate that world rehearsal remains effective as model capacity increases and can scale to stronger backbones while delivering larger performance gains.

%加一个world rehersal随N上升 perfomance上升的分析实验
\begin{table*}[t]
\centering
\scriptsize
\setlength{\tabcolsep}{2.6pt}
\renewcommand{\arraystretch}{0.96}

\resizebox{0.92\textwidth}{!}{%
\begin{tabular}{lcccccccccccc}
\toprule

\multirow{2}{*}{\textbf{Method}}
& \multirow{2}{*}{\textbf{Mode}}
& \multirow{2}{*}{\textbf{Rehearsal}}
& \multicolumn{4}{c}{$\boldsymbol{\tau^2}$\textbf{-bench (\%)}}
& \multicolumn{5}{c}{\textbf{BFCL Multi-Turn (\%)}}
& \multirow{2}{*}{\textbf{Overall (\%)}} \\

\cmidrule(lr){4-7}
\cmidrule(lr){8-12}

& &
& Airline
& Retail
& Telecom
& Avg.
& Base
& Miss Func.
& Miss Param.
& Long Ctx.
& Avg.
& \\

\midrule

Non-TTS
& --
& --
& \second{38.0}
& 42.3
& \second{14.0}
& 31.4
& 47.5
& 45.5
& \best{35.0}
& \second{39.5}
& 41.9
& 36.7 \\

\midrule

TTS
& Par.
& Base
& \second{38.0}
& 45.5
& 13.2
& 32.2
& 50.5
& 46.0
& 30.0
& 39.2
& 41.4
& 36.8 \\

\rowcolor{warmbg}
TTS
& Par.
& \name
& \best{42.0}
& \best{54.4}
& \best{17.5}
& \best{38.0}
& \best{55.5}
& \second{49.0}
& \second{31.0}
& \best{40.0}
& \best{43.9}
& \best{40.9} \\

\midrule

TTS
& Seq.
& Base
& 36.0
& 49.1
& 8.0
& 31.0
& 48.0
& 45.0
& 26.0
& 36.0
& 38.8
& 34.9 \\

\rowcolor{warmbg}
TTS
& Seq.
& \name
& 36.0
& \second{50.9}
& \best{17.5}
& \second{34.8}
& \second{51.5}
& \best{50.5}
& 30.0
& 37.0
& \second{42.3}
& \second{38.5} \\

\bottomrule
\end{tabular}%
}

\caption{
Test-time scaling with $N=2$ on $\tau^2$ and BFCL multi-turn.
Overall is the arithmetic mean of the $\tau^2$-Bench Avg. and BFCL Multi-Turn Avg.
\bestlegend{} Blue cells indicate the highest result in each column,
while  \secondlegend{} red cells indicate the second-highest result.
}
\label{tab:tts_scaling_results}
\end{table*}

\subsection{Test-Time Scaling with World Rehearsal}

% Table~\ref{tab:artis_n2_selected_results} evaluates test-time scaling with two rehearsal attempts under parallel and sequential modes. For each mode, rehearsal is performed with either the base model or the shared \name policy. The parallel Share configuration achieves the best Overall score of 41.0\%, improving the Non-TTS result of 36.7\% by 4.3\%. The improvement is consistent across both benchmarks: the average score increases from 31.4\% to 38.0\% on $\tau^2$-Bench and from 41.9\% to 43.9\% on BFCL Multi-Turn. Sequential Share also improves the Overall score to 38.6\%, showing that world rehearsal benefits final execution under both modes.

% The model used during rehearsal is critical. In both modes, the shared policy consistently outperforms the base model. The base model yields only a marginal improvement under the parallel mode and no improvement under the sequential mode, indicating that the gains do not arise from additional inference compute alone but depend on the environment-response knowledge internalized during training. These results demonstrate that world rehearsal provides an effective test-time scaling mechanism without additional interaction with the external environment.

\paragraph{World Rehearsal Improves Test-Time Performance}
Table~\ref{tab:tts_scaling_results} reports a representative comparison with the rehearsal budget fixed to $N=2$. We evaluate both parallel and sequential modes, using either the base model or \name for rehearsal. Parallel rehearsal with \name achieves the best Overall score of 40.9\%, improving the Non-TTS result of 36.7\% by 4.2\%. The improvement is consistent across both benchmarks: the average score increases from 31.4\% to 38.0\% on $\tau^2$-Bench and from 41.9\% to 43.9\% on BFCL Multi-Turn. Sequential rehearsal with \name also improves the Overall score to 38.5\%, showing that world rehearsal benefits final execution under both modes.

The policy used for rehearsal is important. In both modes, using \name for rehearsal consistently outperforms using the base model. The base model yields only a marginal improvement under the parallel mode and underperforms Non-TTS under the sequential mode. This comparison suggests that the gains do not arise from additional inference compute alone, but depend on the environment-response knowledge internalized through world rehearsal. These results demonstrate that world rehearsal provides an effective mechanism for test-time scaling without additional interaction with the external environment.

\paragraph{Effect of Rehearsal Budget}
We further study the effect of the rehearsal budget on BFCL Multi-Turn performance in Figure~\ref{fig:bfcl-tts-scaling}. As the number of rehearsal attempts increases, the overall performance shows an upward trend. Parallel rehearsal with EnvACE exhibits stronger scaling gains, with the average score increasing from 42.12\% at $N=1$ to 49.50\% at $N=5$; sequential rehearsal also improves from 40.12\% to 45.25\%. Overall, rehearsal with EnvACE achieves higher performance than rehearsal with the base model, suggesting that more accurate environment rehearsal provides more useful information for the final execution. These results further show that increasing the rehearsal budget improves test-time performance, while higher-quality world rehearsal enables more effective use of additional test-time computation. Similar gains are observed on $\tau^2$-Bench, as shown in Figure~\ref{fig:tau2_tts_budget}.

\section{Conclusion}

We introduced \name, an agentic RL method built around world rehearsal. By assigning acting and rehearsal to a single shared policy and optimizing both roles with role-wise GRPO, \name learns not only to select actions but also to generate the corresponding environment responses. Through repeated rehearsal, this knowledge is absorbed into the policy parameters, enabling the policy to internalize environment dynamics as an agent world model and use them to guide subsequent decisions. Across BFCL V4, $\tau^2$-Bench, VitaBench, and FinMCP-Bench, \name achieves strong and transferable tool-use performance, outperforming existing environment-scaling baselines in the overall evaluation. Controlled analyses further demonstrate the effectiveness of world rehearsal, the importance of sharing parameters between acting and rehearsal, and the generalization of these benefits across model scales. Finally, the resulting agent world model enables test-time scaling through private rehearsal, improving final execution without additional interaction with the external environment. Together, these results establish world rehearsal as an effective and scalable alternative to agent training based on real-environment interaction or separate external simulators.

\section{Limitation}

Due to computational constraints, we evaluate \name only up to the 8B scale, leaving its effectiveness on larger models for future investigation. In addition, our current evaluation focuses primarily on tool-interactive tasks, and extending world rehearsal to a broader range of agentic settings remains an interesting direction for future work.

\subsection*{AI use statement}

In this work, we used generative AI tools solely for writing assistance. Specifically, we used large language model based assistants to polish the language of the manuscript, including grammar correction, improving the clarity and conciseness of sentences we had already drafted, and suggesting alternative phrasings for titles and section headings.

We did not use generative AI tools for research ideation, the design of the proposed method, experimental design, data analysis, or the generation of scientific claims or results; these were carried out by the authors. No AI-generated text was incorporated into the manuscript without being reviewed, rewritten where necessary, and verified by the authors. All numbers, tables, figures, and experimental results were produced by our own experimental pipeline. Generative AI tools were not used in the review or evaluation of other researchers' work in connection with this submission, and they are not part of the proposed method's contribution beyond the language models explicitly described as trained policies and baselines in the paper.

\subsection*{Ethics statement}

This work trains and evaluates language model agents for tool use on publicly available
models and benchmarks, and does not involve human subjects, personally identifiable
information, or the release of new user data. All benchmarks we use ($\tau^2$-Bench,
VitaBench, BFCL-v4, and FinMCP-Bench) are existing public benchmarks used under their
intended terms, and their simulated service environments contain synthetic rather than
real customer records. Because our method lets a policy rehearse environment responses
internally, a rehearsed response is a model prediction and not a verified fact about the
world; we therefore emphasize that rehearsal is used for training and for pre-execution
deliberation, and that any deployment in a consequential domain (for example the financial
tool-use setting we evaluate) should keep real execution and verification in the loop.
We see no other ethical concerns specific to this work.

\subsection*{Reproducibility statement}

We describe the world rehearsal formulation and the role-wise optimization objective in
Section~\ref{sec:method} with the complete training procedure, and report the datasets,
backbone models, baselines, and evaluation protocol of every benchmark in
Section~\ref{sec:setup}. All benchmarks are public and we follow each benchmark's official
evaluation protocol without modification, so that reported numbers are directly comparable
to previously published results. Full training and evaluation hyperparameters, including
the learning rate, rollout configuration, sequence lengths, and the maximum number of
interaction turns, are reported in Section~\ref{sec:setup}, together with qualitative
rehearsal trajectories in Appendix~\ref{sec:case_study} that illustrate the exact prompt
format used for the actor and the environment roles. Our code, prompts, and
configuration files are released at the anonymized repository linked in the abstract.

\bibliography{iclr2027_conference}

@misc{cm2,
      title={CM2: Reinforcement Learning with Checklist Rewards for Multi-Turn and Multi-Step Agentic Tool Use}, 
      author={Zhen Zhang and Kaiqiang Song and Xun Wang and Yebowen Hu and Weixiang Yan and others},
      year={2026},
      eprint={2602.12268},
      archivePrefix={arXiv},
      primaryClass={cs.AI},
      url={https://arxiv.org/abs/2602.12268}, 
}

@article{simia,
  title={Simulating environments with reasoning models for agent training},
  author={Li, Yuetai and Inan, Huseyin A and Yue, Xiang and Chen, Wei-Ning and others},
  journal={arXiv preprint arXiv:2511.01824},
  year={2025}
}

@article{dong2026agent,
  title={Agent-world: Scaling real-world environment synthesis for evolving general agent intelligence},
  author={Dong, Guanting and Lu, Junting and Huang, Junjie and Zhong, Wanjun and Liu, Longxiang and Huang, Shijue and Li, Zhenyu and Zhao, Yang and Song, Xiaoshuai and Li, Xiaoxi and others},
  journal={arXiv preprint arXiv:2604.18292},
  year={2026}
}

@article{chen2026vitabench,
  title={VitaBench 2.0: Evaluating Personalized and Proactive Agents in Long-Term User Interactions},
  author={Chen, Yuxin and Zhang, Yi and Cai, Zhengzhou and Shi, Yaorui and Yao, Zhiyuan and Cui, Chenhang and Zheng, Jingnan and Huo, Yaqi and Su, Xi and Gu, Qi and others},
  journal={arXiv preprint arXiv:2605.27141},
  year={2026}
}

@article{noiserl,
  title={Learning to Act under Noise: Enhancing Agent Robustness via Noisy Environments},
  author={Chen, Yuxin and Cai, Xiaodong and Fang, Junfeng and Han, Zhuowen and Wang, Yu and Shi, Yaorui and Zhang, Yi and Gu, Qi and Cai, Xunliang and Wang, Xiang and others},
  journal={arXiv preprint arXiv:2605.27209},
  year={2026}
}

@article{tu2026scaleenv,
  title={ScaleEnv: Scaling Environment Synthesis from Scratch for Generalist Interactive Tool-Use Agent Training},
  author={Tu, Dunwei and Hao, Hongyan and Yang, Hansi and Chen, Yihao and Zhang, Yi-Kai and Xia, Zhikang and Yang, Yu and Sun, Yueqing and Liu, Xingchen and Shen, Furao and others},
  journal={arXiv preprint arXiv:2602.06820},
  year={2026}
}

@article{song2026envscaler,
  title={Envscaler: Scaling tool-interactive environments for llm agent via programmatic synthesis},
  author={Song, Xiaoshuai and Chang, Haofei and Dong, Guanting and Zhu, Yutao and Wen, Ji-Rong and Dou, Zhicheng},
  journal={arXiv preprint arXiv:2601.05808},
  year={2026}
}

@article{lu2026policy,
  title={Policy and World Modeling Co-Training for Language Agents},
  author={Lu, Ning and Lin, Baijiong and Liu, Shengcai and Wu, Jiahao and Lv, Haoze and Wei, Yanbin and Zhu, Lingting and Qian, Shengju and Wang, Xin and Chen, Ying-Cong and others},
  journal={arXiv preprint arXiv:2606.02388},
  year={2026}
}

@article{guo2025deepseekr1,
  title   = {DeepSeek-R1: Incentivizing Reasoning Capability in LLMs via Reinforcement Learning},
  author  = {Guo, Daya and others},
  journal = {arXiv preprint arXiv:2501.12948},
  year    = {2025}
}

@article{xie2024osworld,
  title={Osworld: Benchmarking multimodal agents for open-ended tasks in real computer environments},
  author={Xie, Tianbao and Zhang, Danyang and Chen, Jixuan and Li, Xiaochuan and Zhao, Siheng and Cao, Ruisheng and Hua, Toh J and Cheng, Zhoujun and Shin, Dongchan and Lei, Fangyu and others},
  journal={Advances in Neural Information Processing Systems},
  year={2024}
}

@inproceedings{jimenez2024swe,
  title={Swe-bench: Can language models resolve real-world github issues?},
  author={Jimenez, Carlos E and Yang, John and Wettig, Alexander and Yao, Shunyu and Pei, Kexin and Press, Ofir and Narasimhan, Karthik},
  booktitle={International Conference on Learning Representations},
  year={2024}
}

@article{lambert2024tulu,
  title={Tulu 3: Pushing frontiers in open language model post-training},
  author={Lambert, Nathan and Morrison, Jacob and Pyatkin, Valentina and Huang, Shengyi and Ivison, Hamish and Brahman, Faeze and Miranda, Lester James V and Liu, Alisa and Dziri, Nouha and Lyu, Shane and others},
  journal={arXiv preprint arXiv:2411.15124},
  year={2024}
}

@article{gpt5,
  author  = {{OpenAI}},
  title   = {Introducing GPT-5},
  year    = {2025},
  url     = {https://openai.com/index/introducing-gpt-5/}
}

@article{glm5.2,
  author = {{Z.ai}},
  title  = {{GLM-5.2}: Built for Long-Horizon Tasks},
  year   = {2026},
  url    = {https://z.ai/blog/glm-5.2}
}

@misc{kimik3,
      title={Kimi K3: Open Frontier Intelligence}, 
      author={Kimi Team and Tongtong Bai and Yifan Bai and Yiping Bao and M. C. and Jianfeng Cai and Xinyuan Cai and Peizhou Cao and Yuxuan Cao and Ziwei Chai and Y. Charles and others},
      year={2026},
      eprint={2607.24653},
      archivePrefix={arXiv},
      primaryClass={cs.CL},
      url={https://arxiv.org/abs/2607.24653}, 
}

@InProceedings{patil2025bfcl,
  title = 	 {The Berkeley Function Calling Leaderboard ({BFCL}): From Tool Use to Agentic Evaluation of Large Language Models},
  author =       {Patil, Shishir G and Mao, Huanzhi and Yan, Fanjia and Ji, Charlie Cheng-Jie and Suresh, Vishnu and Stoica, Ion and Gonzalez, Joseph E.},
  booktitle = 	 {Proceedings of the 42nd International Conference on Machine Learning},
  pages = 	 {48371--48392},
  year = 	 {2025},
  editor = 	 {Singh, Aarti and Fazel, Maryam and Hsu, Daniel and Lacoste-Julien, Simon and Berkenkamp, Felix and Maharaj, Tegan and Wagstaff, Kiri and Zhu, Jerry},
  volume = 	 {267},
  series = 	 {Proceedings of Machine Learning Research},
  month = 	 {13--19 Jul},
  publisher =    {PMLR},
  url = 	 {https://proceedings.mlr.press/v267/patil25a.html}
}

@misc{barres2025tau2,
      title={$\tau^2$-Bench: Evaluating Conversational Agents in a Dual-Control Environment}, 
      author={Victor Barres and Honghua Dong and Soham Ray and Xujie Si and Karthik Narasimhan},
      year={2025},
      eprint={2506.07982},
      archivePrefix={arXiv},
      primaryClass={cs.AI},
      url={https://arxiv.org/abs/2506.07982}, 
}

@article{he2025vitabench,
  title={Vitabench: Benchmarking llm agents with versatile interactive tasks in real-world applications},
  author={He, Wei and Sun, Yueqing and Hao, Hongyan and Hao, Xueyuan and Xia, Zhikang and Gu, Qi and Han, Chengcheng and Zhao, Dengchang and Su, Hui and Zhang, Kefeng and others},
  journal={arXiv preprint arXiv:2509.26490},
  year={2025}
}

@inproceedings{zhu2026finmcp,
  title={Finmcp-bench: Benchmarking llm agents for real-world financial tool use under the model context protocol},
  author={Zhu, Jie and Tian, Yimin and Li, Boyang and Wu, Kehao and Liang, Zhongzhi and Li, Junhui and Zhang, Xianyin and Guo, Lifan and Chen, Feng and Liu, Yong and others},
  booktitle={ICASSP 2026-2026 IEEE International Conference on Acoustics, Speech and Signal Processing (ICASSP)},
  pages={19782--19786},
  year={2026},
  organization={IEEE}
}

@article{yao2023react,
  title={React: Synergizing reasoning and acting in language models},
  author={Yao, Shunyu and Zhao, Jeffrey and Yu, Dian and Du, Nan and Shafran, Izhak and Narasimhan, Karthik and Cao, Yuan},
  journal={arXiv preprint arXiv:2210.03629},
  year={2022}
}

@article{yao2026coba,
  title={CoBA-RL: Capability-Oriented Budget Allocation for Reinforcement Learning in LLMs},
  author={Yao, Zhiyuan and Zhang, Yi-Kai and Chen, Yuxin and Sun, Yueqing and Xu, Zishan and Yang, Yu and Hu, Tianhao and Gu, Qi and Su, Hui and Cai, Xunliang},
  journal={arXiv preprint arXiv:2602.03048},
  year={2026}
}

@article{jin2025search,
  title={Search-r1: Training llms to reason and leverage search engines with reinforcement learning},
  author={Jin, Bowen and Zeng, Hansi and Yue, Zhenrui and Yoon, Jinsung and Arik, Sercan and Wang, Dong and Zamani, Hamed and Han, Jiawei},
  journal={arXiv preprint arXiv:2503.09516},
  year={2025}
}

@article{feng2025retool,
  title={Retool: Reinforcement learning for strategic tool use in llms},
  author={Feng, Jiazhan and Huang, Shijue and Qu, Xingwei and Zhang, Ge and Qin, Yujia and Zhong, Baoquan and Jiang, Chengquan and Chi, Jinxin and Zhong, Wanjun},
  journal={arXiv preprint arXiv:2504.11536},
  year={2025}
}

@article{li2025torl,
  title={Torl: Scaling tool-integrated rl},
  author={Li, Xuefeng and Zou, Haoyang and Liu, Pengfei},
  journal={arXiv preprint arXiv:2503.23383},
  year={2025}
}

@inproceedings{qi2025webrl,
  author    = {Zehan Qi and Xiao Liu and Iat Long Iong and Hanyu Lai
               and Xueqiao Sun and Jiadai Sun and Xinyue Yang and Yu Yang
               and Shuntian Yao and Wei Xu and Jie Tang and Yuxiao Dong},
  title     = {{WebRL}: Training {LLM} Web Agents via Self-Evolving Online
               Curriculum Reinforcement Learning},
  booktitle = {International Conference on Learning Representations},
  year      = {2025}
}

@inproceedings{feng2026group,
 author = {Feng, Lang and Xue, Zhenghai and Liu, Tingcong and An, Bo},
 booktitle = {Advances in Neural Information Processing Systems},
 doi = {10.52202/085713-1544},
 editor = {D. Belgrave and C. Zhang and H. Lin and R. Pascanu and P. Koniusz and M. Ghassemi and N. Chen},
 pages = {46375--46408},
 publisher = {Curran Associates, Inc.},
 title = {Group-in-Group Policy Optimization for LLM Agent Training},
 url = {https://proceedings.neurips.cc/paper_files/paper/2025/file/420c9f777c0b4f78d515e53cf74d58b2-Paper-Conference.pdf},
 volume = {38, Main Conference},
 year = {2025}
}

@article{feng2026dr,
  title={Dr. mas: Stable reinforcement learning for multi-agent llm systems},
  author={Feng, Lang and Zheng, Longtao and He, Shuo and Zhang, Fuxiang and An, Bo},
  journal={arXiv preprint arXiv:2602.08847},
  year={2026}
}

@inproceedings{lu2025toolsandbox,
  title={Toolsandbox: A stateful, conversational, interactive evaluation benchmark for llm tool use capabilities},
  author={Lu, Jiarui and Holleis, Thomas and Zhang, Yizhe and Aumayer, Bernhard and Nan, Feng and Bai, Haoping and Ma, Shuang and Ma, Shen and Li, Mengyu and Yin, Guoli and others},
  booktitle={Findings of the Association for Computational Linguistics: NAACL 2025},
  pages={1160--1183},
  year={2025}
}

@misc{yao2024tau,
      title={$\tau$-bench: A Benchmark for Tool-Agent-User Interaction in Real-World Domains}, 
      author={Shunyu Yao and Noah Shinn and Pedram Razavi and Karthik Narasimhan},
      year={2024},
      eprint={2406.12045},
      archivePrefix={arXiv},
      primaryClass={cs.AI},
      url={https://arxiv.org/abs/2406.12045}, 
}

@inproceedings{zeng2024agenttuning,
  title={Agenttuning: Enabling generalized agent abilities for llms},
  author={Zeng, Aohan and Liu, Mingdao and Lu, Rui and Wang, Bowen and Liu, Xiao and Dong, Yuxiao and Tang, Jie},
  booktitle={Findings of the Association for Computational Linguistics: ACL 2024},
  pages={3053--3077},
  year={2024}
}

@inproceedings{liu2025toolace,
 author = {Liu, Weiwen and Huang, Xu and Zeng, Xingshan and hao, xinlong and Yu, Shuai and Li, Dexun and Wang, Shuai and Gan, Weinan and Liu, Zhengying and Yu, Yuanqing and WANG, Zezhong and Wang, Yuxian and Ning, Wu and Hou, Yutai and Wang, Bin and Wu, Chuhan and Xinzhi, Wang and Liu, Yong and Wang, Yasheng and Tang, Duyu and Tu, Dandan and Shang, Lifeng and Jiang, Xin and Tang, Ruiming and Lian, Defu and Liu, Qun and Chen, Enhong},
 booktitle = {International Conference on Learning Representations},
 editor = {Y. Yue and A. Garg and N. Peng and F. Sha and R. Yu},
 pages = {41359--41381},
 title = {ToolACE: Winning the Points of LLM Function Calling},
 url = {https://proceedings.iclr.cc/paper_files/paper/2025/file/663865ea167425c6c562cb0b6bcf76c7-Paper-Conference.pdf},
 volume = {2025},
 year = {2025}
}

@inproceedings{qin2023toolllm,
 author = {Qin, Yujia and Liang, Shihao and Ye, Yining and Zhu, Kunlun and Yan, Lan and Lu, Yaxi and Lin, Yankai and Cong, Xin and Tang, Xiangru and Qian, Bill and Zhao, Sihan and Hong, Lauren and Tian, Runchu and Xie, Ruobing and Zhou, Jie and Gerstein, Mark and li, dahai and Liu, Zhiyuan and Sun, Maosong},
 booktitle = {International Conference on Learning Representations},
 editor = {B. Kim and Y. Yue and S. Chaudhuri and K. Fragkiadaki and M. Khan and Y. Sun},
 pages = {9695--9717},
 title = {ToolLLM: Facilitating Large Language Models to Master 16000+ Real-world APIs},
 url = {https://proceedings.iclr.cc/paper_files/paper/2024/file/28e50ee5b72e90b50e7196fde8ea260e-Paper-Conference.pdf},
 volume = {2024},
 year = {2024}
}

@article{xu2025toucan,
  title={Toucan: Synthesizing 1.5 m tool-agentic data from real-world mcp environments},
  author={Xu, Zhangchen and Soria, Adriana Meza and Tan, Shawn and Roy, Anurag and Agrawal, Ashish Sunil and Poovendran, Radha and Panda, Rameswar},
  journal={arXiv preprint arXiv:2510.01179},
  year={2025}
}

@article{wang2026agent,
  title={Agent world model: Infinity synthetic environments for agentic reinforcement learning},
  author={Wang, Zhaoyang and Xu, Canwen and Liu, Boyi and Wang, Yite and Han, Siwei and Yao, Zhewei and Yao, Huaxiu and He, Yuxiong},
  journal={arXiv preprint arXiv:2602.10090},
  year={2026}
}

@article{xu2026envfactory,
  title={EnvFactory: Scaling Tool-Use Agents via Executable Environments Synthesis and Robust RL},
  author={Xu, Minrui and Wang, Zilin and Deng, Mengyi and Li, Zhiwei and Yang, Zhicheng and Zhu, Xiao and Liu, Yinhong and Zhu, Boyu and Huang, Baiyu and Chen, Chao and others},
  journal={arXiv preprint arXiv:2605.18703},
  year={2026}
}

@article{xiao2026webworld,
  title={Webworld: A large-scale world model for web agent training},
  author={Xiao, Zikai and Tu, Jianhong and Zou, Chuhang and Zuo, Yuxin and Li, Zhi and Wang, Peng and Yu, Bowen and Huang, Fei and Lin, Junyang and Liu, Zuozhu},
  journal={arXiv preprint arXiv:2602.14721},
  year={2026}
}

@article{zeng2026artis,
  title={ARTIS: Agentic Risk-Aware Test-Time Scaling via Iterative Simulation},
  author={Zeng, Xingshan and Wang, Lingzhi and Liu, Weiwen and Li, Liangyou and Wang, Yasheng and Shang, Lifeng and Jiang, Xin and Liu, Qun},
  journal={arXiv preprint arXiv:2602.01709},
  year={2026}
}

@inproceedings{chae2025web,
  title={Web agents with world models: Learning and leveraging environment dynamics in web navigation},
  author={Chae, Hyungjoo and Kim, Namyoung and Ong, Kai and Gwak, Minju and Song, Gwanwoo and Kim, Jihoon and Kim, Sunghwan and Lee, Dongha and Yeo, Jinyoung},
  booktitle={International Conference on Learning Representations},
  volume={2025},
  pages={63707--63738},
  year={2025}
}

@article{yu2026reinforcement,
  title={Reinforcement World Model Learning for LLM-based Agents},
  author={Yu, Xiao and Peng, Baolin and Xu, Ruize and Shen, Yelong and He, Pengcheng and Nath, Suman and Singh, Nikhil and Gao, Jiangfeng and Yu, Zhou},
  journal={arXiv preprint arXiv:2602.05842},
  year={2026}
}

@article{wang2026role,
  title={Role-Agent: Bootstrapping LLM Agents via Dual-Role Evolution},
  author={Wang, Xucong and Ma, Ziyu and Yang, Shidong and Huang, Tongwen and Wang, Pengkun and Wang, Yong and Chu, Xiangxiang},
  journal={arXiv preprint arXiv:2606.10917},
  year={2026}
}

@inproceedings{lu2026ui,
  title={Ui-r1: Enhancing efficient action prediction of gui agents by reinforcement learning},
  author={Lu, Zhengxi and Chai, Yuxiang and Guo, Yaxuan and Yin, Xi and Liu, Liang and Wang, Hao and Xiao, Han and Ren, Shuai and Zhao, Pengxiang and Liu, Guangyi and others},
  booktitle={Proceedings of the AAAI Conference on Artificial Intelligence},
  volume={40},
  number={21},
  pages={17608--17616},
  year={2026}
}

@article{singh2025agentic,
  title={Agentic reasoning and tool integration for llms via reinforcement learning},
  author={Singh, Joykirat and Magazine, Raghav and Pandya, Yash and Nambi, Akshay},
  journal={arXiv preprint arXiv:2505.01441},
  year={2025}
}

@inproceedings{wei2026swe,
 author = {Wei, Yuxiang and Duchenne, Olivier and Copet, Jade and Carbonneaux, Quentin and ZHANG, LINGMING and Fried, Daniel and Synnaeve, Gabriel and Singh, Rishabh and Wang, Sida},
 booktitle = {Advances in Neural Information Processing Systems},
 doi = {10.52202/085713-2629},
 editor = {D. Belgrave and C. Zhang and H. Lin and R. Pascanu and P. Koniusz and M. Ghassemi and N. Chen},
 pages = {78500--78525},
 publisher = {Curran Associates, Inc.},
 title = {SWE-RL: Advancing LLM Reasoning via Reinforcement Learning on Open Software Evolution},
 url = {https://proceedings.neurips.cc/paper_files/paper/2025/file/7107d4d2e837bde2171c6b71b5bde954-Paper-Conference.pdf},
 volume = {38, Main Conference},
 year = {2025}
}

@article{yang2025qwen3,
  title={Qwen3 technical report},
  author={Yang, An and Li, Anfeng and Yang, Baosong and Zhang, Beichen and Hui, Binyuan and Zheng, Bo and Yu, Bowen and Gao, Chang and Huang, Chengen and Lv, Chenxu and others},
  journal={arXiv preprint arXiv:2505.09388},
  year={2025}
}

@article{guo2025world,
  title={World modelling improves language model agents},
  author={Guo, Shangmin and Domingues, Omar Darwiche and Avalos, Rapha{\"e}l and Courville, Aaron and Strub, Florian},
  journal={arXiv preprint arXiv:2506.02918},
  year={2025}
}

@inproceedings{ruan2024identifying,
  title={Identifying the risks of lm agents with an lm-emulated sandbox},
  author={Ruan, Yangjun and Dong, Honghua and Wang, Andrew and Pitis, Silviu and Zhou, Yongchao and Ba, Jimmy and Dubois, Yann and Maddison, Chris and Hashimoto, Tatsunori},
  booktitle={International Conference on Learning Representations},
  volume={2024},
  pages={27031--27098},
  year={2024}
}

@article{zuo2026qwen,
  title={Qwen-AgentWorld: Language World Models for General Agents},
  author={Zuo, Yuxin and Xiao, Zikai and Sheng, Li and Huang, Fei and Tu, Jianhong and Liu, Yuxuan and Tang, Tianyi and Hu, Xiaomeng and Su, Yang and Lan, Qingfeng and others},
  journal={arXiv preprint arXiv:2606.24597},
  year={2026}
}

@article{chen2025internalizing,
  title={Internalizing world models via self-play finetuning for agentic rl},
  author={Chen, Shiqi and Zhu, Tongyao and Wang, Zian and Zhang, Jinghan and Wang, Kangrui and Gao, Siyang and Xiao, Teng and Teh, Yee Whye and He, Junxian and Li, Manling},
  journal={arXiv preprint arXiv:2510.15047},
  year={2025}
}

@article{liu2026comap,
  title={COMAP: Co-Evolving World Models and Agent Policies for LLM Agents},
  author={Liu, Youwei and Wang, Jian and Wang, Hanlin and Li, Wenjie},
  journal={arXiv preprint arXiv:2606.02372},
  year={2026}
}

@article{cai2026beyond,
  title={Beyond Next-Observation Prediction: Agent-Authored World Modeling for Sequential Decision Making},
  author={Cai, Guangfeng and Yang, Kaibing and He, Shuo and Li, Yu and Yang, Shengtian and Lv, Jiaqi and Feng, Lei},
  journal={arXiv preprint arXiv:2606.25421},
  year={2026}
}

@article{lu2026sdar,
  title={Self-distilled agentic reinforcement learning},
  author={Lu, Zhengxi and Yao, Zhiyuan and Han, Zhuowen and Wang, Zi-Han and Wu, Jinyang and Gu, Qi and Cai, Xunliang and Lu, Weiming and Xiao, Jun and Zhuang, Yueting and others},
  journal={arXiv preprint arXiv:2605.15155},
  year={2026}
}
\bibliographystyle{iclr2027_conference}

\clearpage
\appendix
% \section{Appendix}
\section{Case Study}
\label{sec:case_study}

\begin{figure*}[htbp]
\centering
\includegraphics[width=\textwidth]{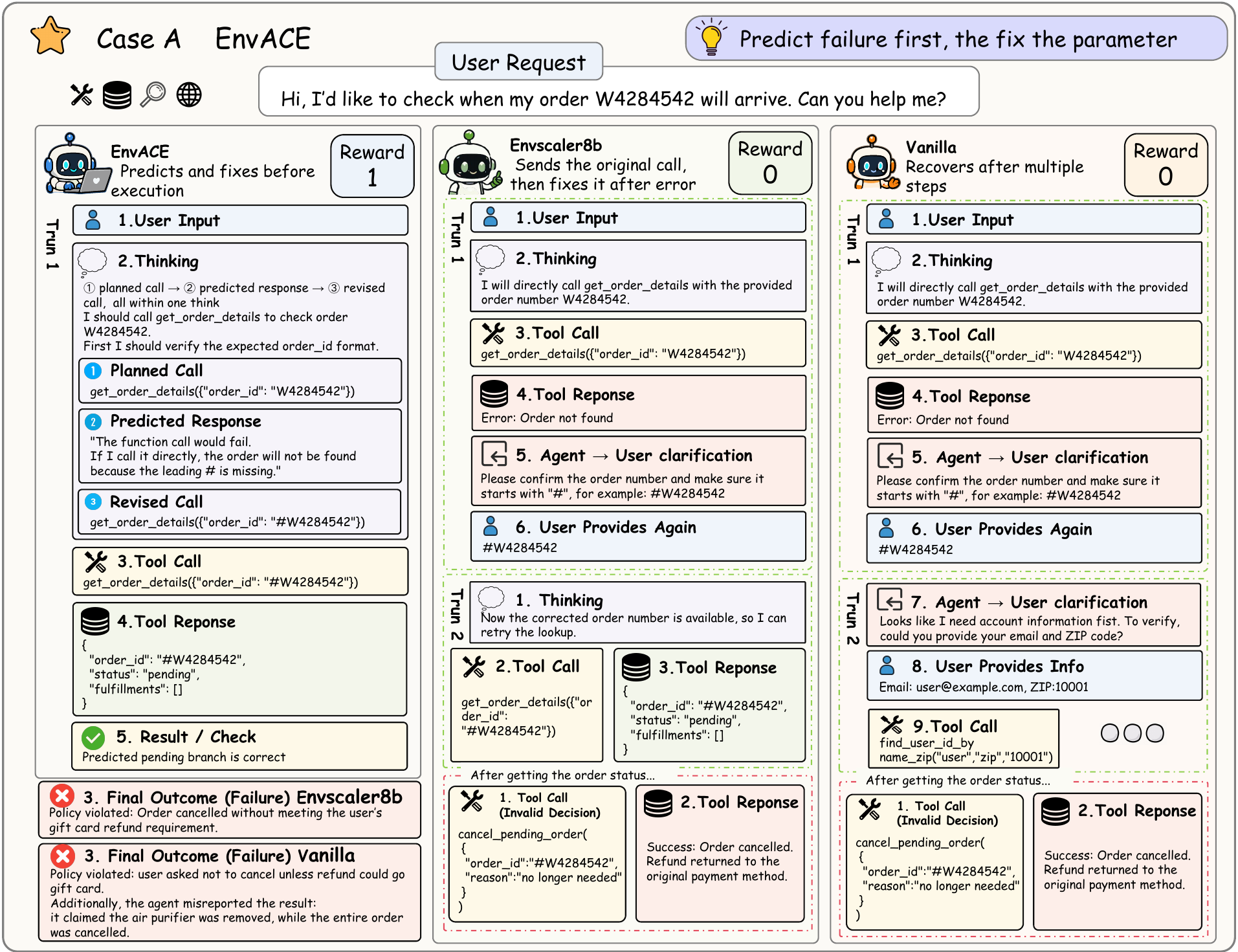}
\caption{
  Case study of \name compared with EnvScaler8B and Vanilla agents.
  \name predicts potential tool-call failures before execution and
  automatically repairs invalid parameters, while baseline agents require
  additional recovery steps after failure.
}
\label{fig:envace_case}
\end{figure*}

Figure~\ref{fig:envace_case} presents a representative example comparing
\name with EnvScaler8B and Vanilla agents. In this example, \name anticipates
a potential tool-call failure and repairs the invalid parameters before
execution. In contrast, the baseline agents issue an invalid tool call and
require additional interaction steps to recover from the resulting failure.
This case illustrates how world rehearsal enables more reliable and efficient
tool use.

\begin{figure*}[htbp]
\centering
\includegraphics[width=\textwidth]{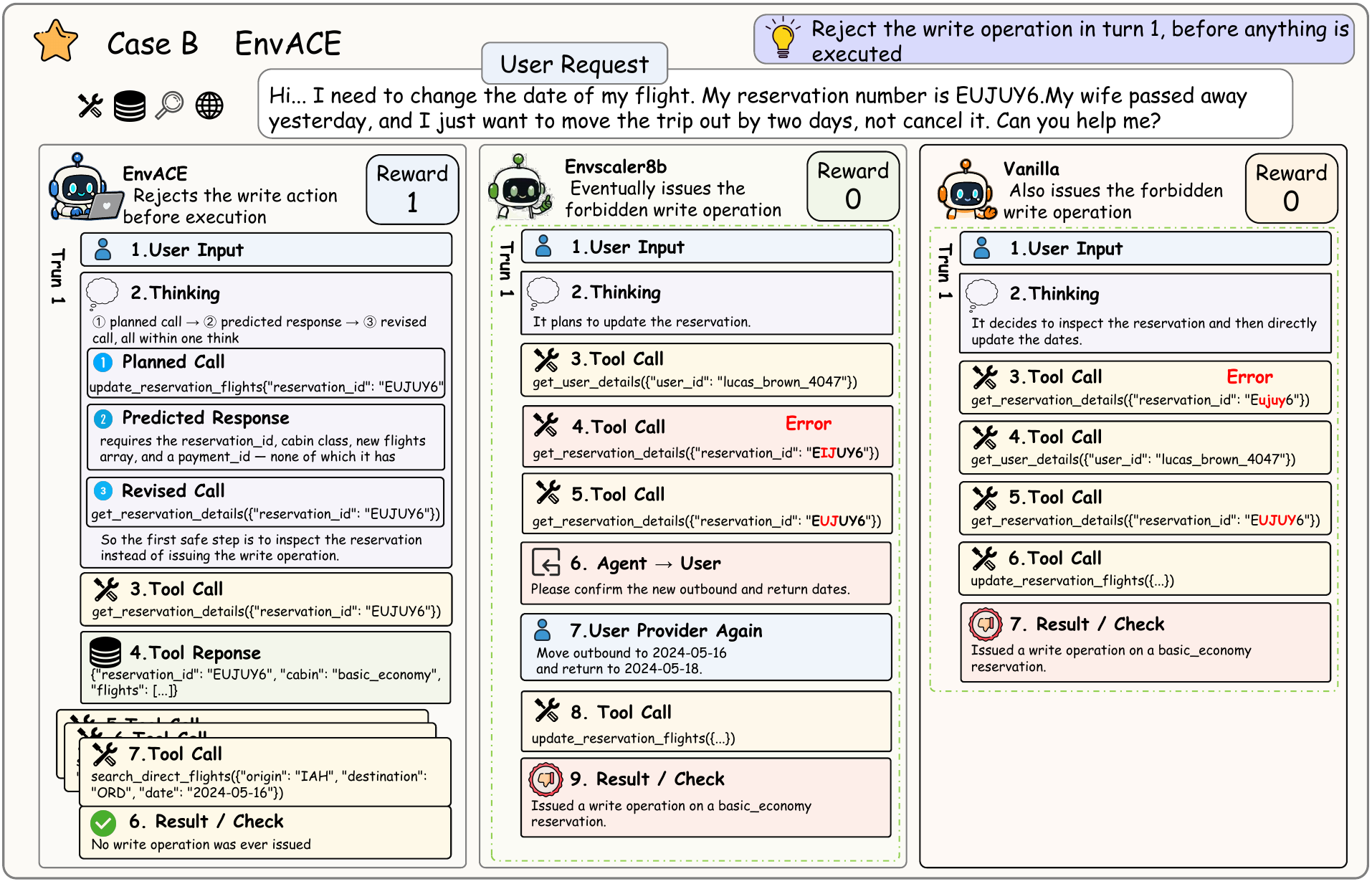}
\caption{
  Case study of \name compared with EnvScaler8B and Vanilla agents.
  \name performs world rehearsal before tool execution, predicting that the planned
  write operation would fail and proactively replacing it with a safe read-only query.
  In contrast, baseline agents fail to anticipate the consequence of tool calls and
  eventually execute invalid write operations.
}
\label{fig:envace_case2}
\end{figure*}

\clearpage
As shown in Figure~\ref{fig:envace_case2}, this case illustrates how \name avoids an
invalid write operation through world rehearsal. The user requests to modify a flight
reservation with a basic economy ticket. Although updating the reservation appears to be
a reasonable next step, such a write operation is not allowed under the given environment
constraints. Before execution, \name performs internal world rehearsal by simulating the
planned tool call and predicting its potential response. It recognizes that the required
information for executing \texttt{update\_reservation\_flights} is incomplete and that
issuing the write operation would lead to an invalid action. Therefore, \name proactively
replaces the unsafe action with a read-only query to obtain reservation details, avoiding
any forbidden modification. In contrast, EnvScaler8B and Vanilla agents fail to anticipate
the consequence of tool calls and eventually execute the invalid write operation, resulting
in failure. This example demonstrates that world rehearsal enables \name to anticipate tool
consequences and make safer decisions before interacting with the external environment.
% \subsection{Training and Evaluation Details}
% \label{sec:Training and Evaluation Details}
% \paragraph{Training Hyperparameters.}
% Table~\ref{tab:training_hyperparameters} summarizes the hyperparameters used for our reinforcement learning training.

% \begin{table}[h]
% \centering
% \caption{Training hyperparameters.}
% \label{tab:training_hyperparameters}
% \begin{tabular}{lc}
% \toprule
% Hyperparameter & Value \\
% \midrule
% \multicolumn{2}{c}{Optimization} \\
% \midrule
% Learning rate & $1\times10^{-6}$ \\
% Batch size & 16 \\
% Rollouts per task ($G$) & 4 \\
% Instances per step & 64 \\
% Max optimization steps & 470 \\
% KL coefficient & $1\times10^{-4}$ \\
% Entropy coefficient & 0.0 \\
% \midrule
% \multicolumn{2}{c}{Rollout} \\
% \midrule
% Temperature & 1.0 \\
% Max input length & 12,000 \\
% Max response length & 8,000 \\
% \midrule
% \multicolumn{2}{c}{Agent} \\
% \midrule
% Max interaction turns & 30 \\
% \bottomrule
% \end{tabular}
% \end{table}
\clearpage
\section{Test-Time Scaling Procedure}
\label{sec:tts_details}
Algorithm~\ref{alg:test_time_scaling}  summarizes the \emph{rehearse--evaluate--summarize} procedure used for test-time scaling. Given a task $x$, the policy $\pi$ performs $N$ private rehearsal attempts before interacting with the external environment. In the parallel mode, attempts are generated independently, whereas in the sequential mode, each attempt conditions on previous rehearsal trajectories and feedback for iterative refinement. All rehearsal results are summarized into a compact rehearsal memory $m_x$, which guides a single committed execution in the real environment. Only this final execution is officially scored. In our experiments, the acting policy is EnvACE-8B, and the rehearsal model $M$ is either Qwen3-8B or EnvACE-8B. 

\begin{algorithm}[h]
\caption{Test-Time Scaling with World Rehearsal}
\label{alg:test_time_scaling}
\begin{algorithmic}[1]
\Require Task $x$, acting policy $\pi$, rehearsal model $M$,
rehearsal budget $N$, mode
$m \in \{\textsc{Parallel}, \textsc{Sequential}\}$
\Ensure Final trajectory $\tau^\star$ and official score $r$

\State Initialize rehearsal history $\mathcal{H} \gets [\,]$

\For{$n = 1,\ldots,N$}
    \State $c^{(n)} \gets
    \begin{cases}
    \emptyset,
    & m=\textsc{Parallel},\\
    \{(\tilde{\tau}^{(j)},f^{(j)})\}_{j<n},
    & m=\textsc{Sequential}
    \end{cases}$

    \State $\tilde{\tau}^{(n)}
    \gets \Call{Rehearse}{\pi,M,x,c^{(n)}}$

    \State $f^{(n)}
    \gets \Call{SelfEvaluate}{\pi,x,\tilde{\tau}^{(n)}}$

    \State $\mathcal{H}
    \gets
    \mathcal{H}\cup
    \{(\tilde{\tau}^{(n)},f^{(n)})\}$
\EndFor

\State $m_x
\gets \Call{Summarize}{\pi,x,\mathcal{H}}$

\State Initialize $\mathcal{E}_{\mathrm{real}}$
from the task's initial context

\State $\tau^\star
\gets
\Call{Execute}{\pi,x,m_x,\mathcal{E}_{\mathrm{real}}}$

\State $r
\gets
\Call{OfficialEvaluate}{x,\tau^\star}$

\State \Return $(\tau^\star,r)$
\end{algorithmic}
\end{algorithm}
\section{Test-Time Scaling on $\tau^2$-Bench.}
\begin{figure}[h]
    \centering
    \includegraphics[width=0.75\linewidth]{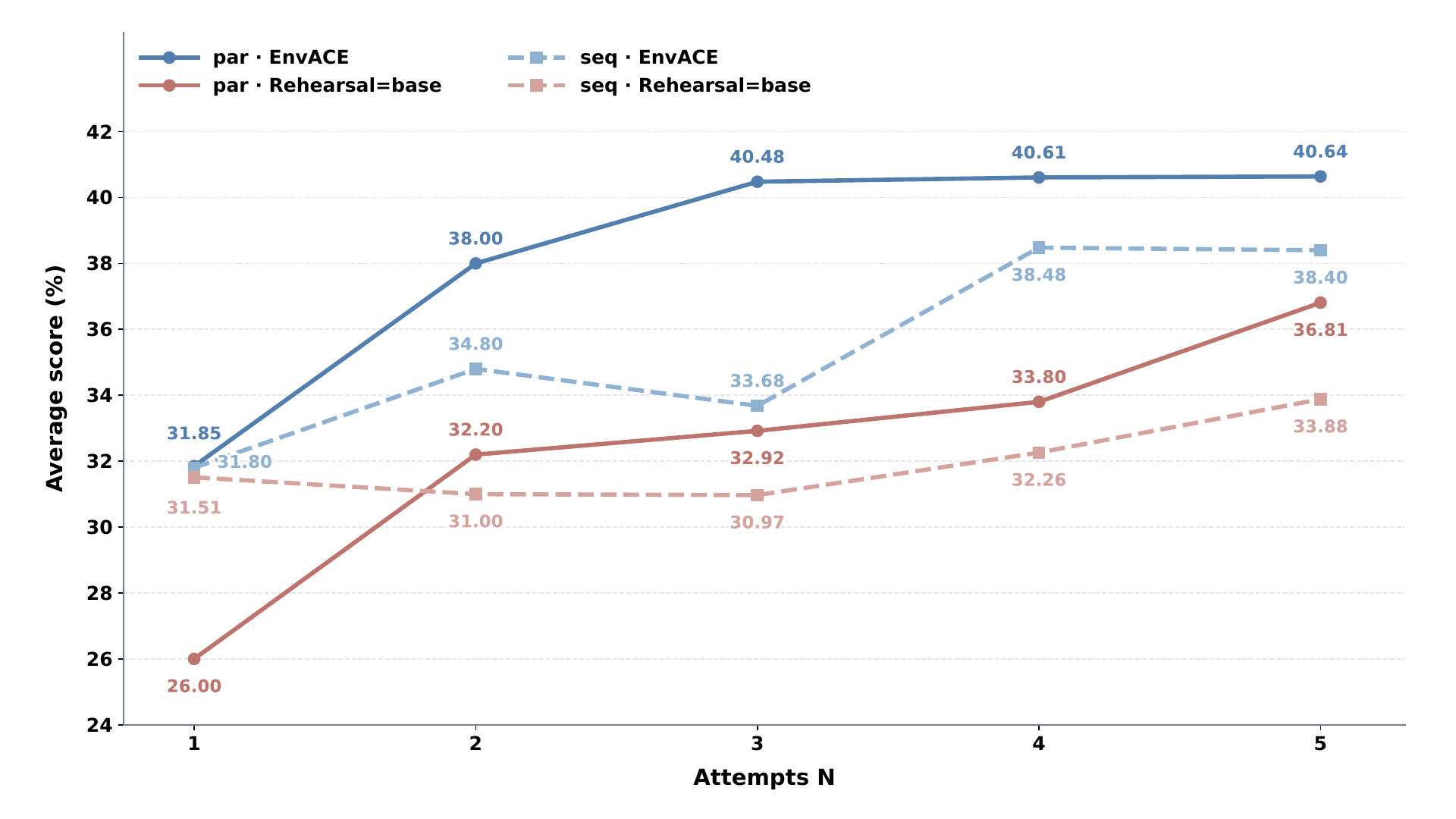}
    \caption{TTS scaling performance on $\tau^2$-Bench.}
    \label{fig:tau2_tts_budget}
\end{figure}

We further evaluate the effect of the rehearsal budget on $\tau^2$-Bench in
Figure~\ref{fig:tau2_tts_budget}. Performance improves overall as the number of
rehearsal attempts increases under both parallel and sequential modes. With
EnvACE rehearsal, the parallel score increases from 31.85\% to 40.64\%, while
the sequential score increases from 31.80\% to 38.40\%. Across all evaluated
budgets, EnvACE outperforms the corresponding base-model rehearsal, indicating
that its internalized environment-response knowledge enables more effective
test-time rehearsal.
\clearpage
\section{Prompts}
\label{sec:prompts}
The following prompt templates instantiate the main components of the
test-time scaling procedure described in Section~\ref{sec:test_time_scaling}
and Appendix~\ref{sec:tts_details}. The world-rehearsal prompt specifies how
environment responses are generated conditioned on the current interaction
context and agent action. The self-evaluation prompt produces an assessment
and revision feedback for each rehearsed trajectory, while the summarization
prompt aggregates the resulting trajectories and feedback into a compact
rehearsal memory. For sequential rehearsal, an additional history template
organizes previous trajectories and feedback as context for subsequent
attempts.

% ===============================
% Color Style
% ===============================

\definecolor{mainblue}{RGB}{65,120,210}
\definecolor{lightblue}{RGB}{240,246,255}
\definecolor{softyellow}{RGB}{255,249,220}
\definecolor{deepyellow}{RGB}{245,195,55}

% ===============================
% Main Appendix Box
% ===============================

% ===============================
% Rehearsal Prompt Box
% ===============================

\newtcolorbox{rehearsalbox}[1]{
    enhanced,
    breakable,
    width=\linewidth,
    colback=lightblue,
    colframe=mainblue,
    boxrule=1.0pt,
    arc=2mm,
    left=5mm,
    right=5mm,
    top=6mm,
    bottom=5mm,
    title={#1},
    fonttitle=\bfseries,
    coltitle=white,
    colbacktitle=mainblue
}

% ===============================
% Code Example Box
% ===============================

\newtcblisting{codebox}{
    enhanced,
    breakable,
    width=\linewidth,
    colback=white,
    colframe=mainblue,
    boxrule=0.7pt,
    arc=1.5mm,
    left=3mm,
    right=3mm,
    top=2mm,
    bottom=2mm,
    listing only,
    listing options={
        basicstyle=\ttfamily\scriptsize,
        breaklines=true,
        breakatwhitespace=false,
        columns=fullflexible,
        keepspaces=true
    }
}

\begin{rehearsalbox}
{World Rehearsal Prompt for Agentic Tool Execution}

You are a precise tool simulator.

Your task: validate the RL model's tool call against the candidate tools
whitelist and the JSON schema, then generate the tool execution result.

\textbf{Input Information}

You will be given:

\begin{itemize}[leftmargin=1.5em]

\item The candidate tools whitelist (list of tool schemas this sample is allowed to call).

\item Previous tool calls and results (few-shot, ground-truth reference data;
may be \texttt{<unavailable>} when no prior calls exist).

\item Current conversation history.

\item The RL model's latest response (containing
\texttt{<tool\_call>...</tool\_call>} tags).

\item The current tool to simulate: name, JSON schema, parsed arguments.

\end{itemize}

\textbf{Validation Procedure}

\textbf{1. Tool Call Format Check}

The RL model's latest response must wrap the tool call inside
\texttt{<tool\_call>...</tool\_call>} tags, with valid JSON:

\[
\{\texttt{"name": "...", "arguments": \{...\}}\}
\]

On failure (tool\_call wrapper absent, or JSON malformed), output:

\begin{codebox}
<execution_result>
{
 "error_tool_call":{
   "code":"INVALID_TOOL_CALL_FORMAT",
   "message":"<concise reason>"
 }
}
</execution_result>
\end{codebox}

\textbf{2. Tool Whitelist Check}

The tool name (shown under "Current tool to simulate") must appear in the
candidate tools whitelist (either the explicit list, or the \texttt{<tools>}
block inside the caller's system prompt).

On failure, output:

\begin{codebox}
<execution_result>
{
 "error_tool_call":{
   "code":"TOOL_NOT_AVAILABLE",
   "message":"tool name '<name>' is not found. Please check your function call and use provided tools."
 }
}
</execution_result>
\end{codebox}

\textbf{3. Argument Schema Check}

Arguments must be a JSON object satisfying the input schema:

\begin{itemize}[leftmargin=1.5em]

\item Required fields must be present.

\item Primitive types must match.

\item Enum values must match.

\end{itemize}

On failure, output:

\begin{codebox}
<execution_result>
{
 "error_tool_call":{
   "code":"INVALID_PARAMETERS",
   "message":"<concise reason>",
   "details":{
      "errors":[...]
   }
 }
}
</execution_result>
\end{codebox}

\textbf{4. Tool Response Generation}

Only when steps 1, 2, and 3 all pass:

\begin{itemize}[leftmargin=1.5em]

\item If a related call appears in the few-shot section with matching name
and similar arguments, reuse its execution result content.

\item Otherwise, generate a reasonable execution result that strictly satisfies
the input schema and stays factually consistent with the few-shot patterns.

\item May need to fix some type or error in the examples.

\end{itemize}

\textbf{Output Rules}

Apply to both success and error paths:

\begin{itemize}[leftmargin=1.5em]

\item Wrap the result in
\texttt{<execution\_result>...</execution\_result>} XML tags.

\item Wrapped content must be valid JSON: a non-empty object or an array.

\item Return ONLY the wrapped JSON; no commentary, no markdown, no code fences.

\item Act as a silent function executor: no explanations, no suggestions,
no hints, no references to examples or comparisons.

\end{itemize}

\textbf{Format:}

\begin{codebox}
<execution_result>
<JSON object or array>
</execution_result>
\end{codebox}

\end{rehearsalbox}
\begin{rehearsalbox}
{Self-Evaluation Prompt}

You are an agentic reasoning supervisor.

Your job is to evaluate the agent's last simulated action in a given
environment and provide actionable suggestions for improvement.

You must be strict, concise, and goal-oriented.
Always assume the agent is allowed to revise its plan.

\textbf{Input Information}

You will be given:

\begin{itemize}[leftmargin=1.5em]

\item The environment description:

\texttt{\{tool\_documents\}}

\item The conversation history:

\texttt{\{history\}}

\item The agent's simulated action and returned result:

\texttt{\{simulation\}}

\end{itemize}

\textbf{Evaluation Procedure}

Your tasks:

\begin{enumerate}[leftmargin=1.5em]

\item Judge whether the agent's action is correct, helpful, and aligned
with the user's task goal.

\item Identify any mistakes or weaknesses in the action.

\item Explain why the action failed or is suboptimal, based strictly on
the environment feedback.

\item Return the exact evaluation result within the
\texttt{<Result>} and \texttt{</Result>} tags:

\begin{itemize}
    \item Output 1 if the action is correct.
    \item Output 0 if the action is incorrect.
\end{itemize}

\item If the action is incorrect, provide a revised action or an improved
action plan for the next simulation attempt.

\end{enumerate}

\textbf{Output Format}

\begin{codebox}
<Evaluation>
(Detailed judgement and explanation)
</Evaluation>

<Result>
1/0
</Result>

<Suggestion>
(Detailed suggestion)
</Suggestion>
\end{codebox}

\end{rehearsalbox}
\begin{rehearsalbox}
{Summarization Prompt for Final Execution Recommendation}

You are a senior execution advisor.

Your job is to summarize multiple simulated attempts and evaluations into
a single, reliable execution recommendation.

You do not simulate tools or evaluate correctness.

You produce a final actionable recommendation that will be executed once
in a real environment.

\textbf{Input Information}

You will receive:

\begin{itemize}[leftmargin=1.5em]

\item The environment description:

\texttt{\{tool\_documents\}}

\item The conversation history:

\texttt{\{history\}}

\item Multiple simulated attempts and feedback:

\texttt{\{Simulation\_history\}}

\end{itemize}

Each simulated attempt may include:

\begin{itemize}[leftmargin=1.5em]

\item The action taken.

\item The simulated tool return.

\item Optional self-evaluation and suggestions.

\end{itemize}

\textbf{Recommendation Procedure}

Your tasks:

\begin{enumerate}[leftmargin=1.5em]

\item Identify the most reliable execution strategy based on all simulated
attempts.

\item Extract key lessons from failures and partial successes.

\item Determine the safest and most likely-to-succeed action plan.

\end{enumerate}

\textbf{Important Rules}

\begin{itemize}[leftmargin=1.5em]

\item Do not repeat the full history.

\item Do not output multiple alternative plans.

\item Produce ONE final recommendation using natural language.

\item Be conservative: prefer actions that are robust and verified in
simulation.

\item You may include a step-by-step action sequence, but clearly state
that it is FOR REFERENCE ONLY (not to be followed blindly) and explain
why each step is taken.

\end{itemize}

\textbf{Output Format}

\begin{codebox}
{
"recommendation": "...",
"rationale": "..."
}
\end{codebox}

\end{rehearsalbox}

\begin{rehearsalbox}
{Previous Attempt History Prompt}

Previous attempts that you can refer to:

\begin{codebox}
<Attempt>
<Action>{action}</Action>
<Evaluation>{evaluation}</Evaluation>
<Suggestion>{suggestion}</Suggestion>
</Attempt>
\end{codebox}

\end{rehearsalbox}

% \subsection{Training and Evaluation Details}
% \label{sec:Training and Evaluation Details}
% \paragraph{Training Hyperparameters.}
% Table~\ref{tab:training_hyperparameters} summarizes the hyperparameters used for our reinforcement learning training.

% \begin{table}[h]
% \centering
% \caption{Training hyperparameters.}
% \label{tab:training_hyperparameters}
% \begin{tabular}{lc}
% \toprule
% Hyperparameter & Value \\
% \midrule
% \multicolumn{2}{c}{Optimization} \\
% \midrule
% Learning rate & $1\times10^{-6}$ \\
% Batch size & 16 \\
% Rollouts per task ($G$) & 4 \\
% Instances per step & 64 \\
% Max optimization steps & 470 \\
% KL coefficient & $1\times10^{-4}$ \\
% Entropy coefficient & 0.0 \\
% \midrule
% \multicolumn{2}{c}{Rollout} \\
% \midrule
% Temperature & 1.0 \\
% Max input length & 12,000 \\
% Max response length & 8,000 \\
% \midrule
% \multicolumn{2}{c}{Agent} \\
% \midrule
% Max interaction turns & 30 \\
% \bottomrule
% \end{tabular}
% \end{table}

\end{document}